\documentclass[onecolumn,3p]{elsarticle}

\usepackage{url}
\usepackage{amssymb}
\usepackage{graphicx,graphics}
\usepackage[usenames]{color}
\usepackage{natbib}
\usepackage{amsmath}
\usepackage{subfig}
\usepackage[nomarkers,tablesonly,notablist]{endfloat}
\definecolor{MyBackground}{rgb}{1,0.7,0.7} 
\definecolor{MyForeground}{rgb}{0,0,0}

\DeclareMathOperator{\avg}{avg}
\setcitestyle{authoryear}
\setcitestyle{round}

\begin{document}

\begin{frontmatter}


 %\title{Title}\thanksref{label1}}

\title{Explainable Deep Learning for Augmentation of sRNA Expression Profiles}

%\author{Alexander Andronov\corauthref{cor1}\thanksref{label2}}

%\author[label1,label2]{Alexander Andronov, Helen Fioshina, Maxim Fioshin }
%\address[label1]{}
%\address[label2]{}

\author[lab1]{Jelena Fiosina}
\author[lab2,lab3,lab4]{Maksims Fiosins}
\author[lab2,lab3]{Stefan Bonn}

\address[lab1]{Clausthal University of Technology, Clausthal-Zellerfeld}
\address[lab2]{German Center for Neurodegenerative Diseases, T\"ubingen}
\address[lab3]{Institute for Medical Systems Biology, 
University Medical Center Hamburg-Eppendorf, Hamburg}
\address[lab4]{Genevention GmbH, G\"ottingen, Germany}

\begin{abstract}
The lack of well-structured metadata annotations complicates the  re-usability and interpretation of the growing amount of publicly available RNA expression data. The machine learning-based prediction of metadata (data augmentation) can considerably improve the quality of expression data annotation. In this study, we systematically benchmark deep learning (DL) and random forest (RF)-based metadata augmentation of tissue, age, and sex using small RNA (sRNA) expression profiles. We use 4243 annotated sRNA-Seq samples from the small RNA expression atlas (SEA) database to train and test the augmentation performance. In general, the DL machine learner outperforms the RF method in almost all tested cases. The average cross-validated prediction accuracy of the DL algorithm for tissues is 96.5\%, for sex is 77\%, and for age is 77.2\%. The average tissue prediction accuracy for a completely new dataset is 83.1\% (DL) and 80.8\% (RF). To understand which sRNAs influence DL predictions, we employ backpropagation-based feature importance scores using the DeepLIFT method, which enable us to obtain information on biological relevnace of sRNAs. 
\end{abstract}

%, and considerably improves classification performance for 'unseen' datasets.

\begin{keyword}
augmentation \sep classification \sep deep learning \sep explainable artificial intelligence \sep random forest \sep ontology \sep  small RNA expression \sep contamination
\end{keyword}
\end{frontmatter}

\section{Introduction}
\label{sec1}

Data annotations (tissue, age, sex, etc.) are crucial for the re-use of data. A detailed description of the biological conditions in which data has been obtained is required to extract new information from the obtained data. The data should be findable, accessible, interoperable, and reusable, which ultimately facilitates knowledge discovery \citep{Wilkinson16}. Annotations are an essential part of semantic data integration systems and allow for a deeper analysis of the data \citep{Madan18}. So far, metadata is often not stored together with the expression data and the available metadata is often not normalized, and is unstructured and incomplete. The widely used GEO repository \citep{GEO}, for example, stores annotations as a number of free-text description fields. This leads to missing and/or inaccurate annotations and requires revisions and manual corrections by an expert \citep{Hadley18}.
In this study, we aim to predict the metadata based on deep-sequenced small RNAs' (sRNAs') expression profiles by formulating this prediction as a classification problem. sRNAs are short (less then 200 nt), usually non-coding RNA molecules with many crucial biological functions \citep{Storz1260}. The basic rationale for this approach is that data with similar sRNA expressions should have similar metadata. Based on this assumption, we hypothesize that sRNA expression profiles contain enough information to predict the sRNA tissue, age, and sex accurately. We believe that deep learning (DL)-based algorithms might outperform more conventional random forest (RF)-based machine learners (MLs) in sRNA metadata prediction, if enough training data is available. We also hypothesize that backpropagation-based feature importance scores may help to biologically rationalize the classification process of DL.

To distinguish between biological conditions, different ML methods were applied. In \citep{Guo17} and \citep{Hadley18}, the sex in different micro RNA (miRNA) tissue samples was defined using differential expression (DE) analysis. In \citep{Hadley18}, the authors used DE analysis and analysis of variance to detect the sex differences in several tissues in miRNAs. 
In \citep{Ellis18}, the age, sex, and tissue were predicted from mRNA sequencing (mRNA-Seq) expression data using a regression-based approach. massiR \citep{Buckberry14} is a method for sex prediction based on gene expression microarrays using clustering.
%In \citep{Simon14}, the sex based on mRNA expression was predicted, and the most important mRNAs were selected. 
Many studies use an RF method for the classification of expression data, particularly in disease diagnostics \citep{Statnikov08}. \citep{Johnson18} provides a good overview of ML methods for expression data analysis.

For our analysis, we used data from the sRNA expression atlas (SEA, http://sea.ims.bio) \citep{Rahman17}, a database containing well-structured, manually curated, ontology-based annotations of publicly available sRNA-Seq data. All data from the SEA was analyzed with the same workflow
(OASIS \citep{Rahman18}, https://oasis.dzne.de). We used 4243 annotated human sRNA-Seq samples from the SEA.

We applied the DL and RF classifiers for the considered augmentation problem and compared their results. 
The RF classsifier is an ensemble-based classifier, which outperforms other
conventional classifiers for very high-dimensional data \citep{Breiman2001}. An RF classifier requires lesser training data in comparison with the DL classifier and allows the interpretation of features by generating variable importances. However, the RF classifier is sensitive to class imbalance \citep{obrein19}.

%Deep learning (DL) has introduced major advances for solving problems that have remained unsolved by the artificial intelligence community for many years \citep{Lecun15}. 
DL is able to analyze big data and is robust enough to treat large amounts of noisy training data \citep{Lecun15}, \citep{Xiao15}. Its disadvantage is that, it requires large amounts of training data \citep{Li19}, is prone to overfit for small training sets and is difficult to biologically interpret (feature importance) \citep{Webb18}. In \citep{Kong18} the RF and DL approaches were used in two stages. For the first stage, the RF approach was used to extract the most important features and then for the second stage, the DL approach was implemented for gene expression data classification based on the selected features.
Many researchers are currently trying to explain DL models \citep{Bach15}, \citep{Montavon17}, \citep{NIPS2016_6321}. Some methods are model-agnostic, which can explain the behavior of every "black box" or "grey box" model: \citep{Lakkaraju17}, \citep{Ribeiro16}, \citep{Molnar19}. Some methods are model specific, such as perturbation-based \citep{Robnik18} or backpropagation-based \citep{Shrikumar17} models. We have used DeepLIFT \citep{Shrikumar17} scores to explain DL models. 

In this study, we present that DL algorithms outperform RF-based data augmentation for tissue, sex, and age annotations using sRNA expression profiles, if enough training data is available. 
More specifically, the DL method performs better than the RF method for cross-validation experiments as well as on "one dataset out" experiments. We have demonstrated how backpropagation can provide a biological interpretation of relevant features for the DL classification. 

\section{Methods}
\subsection{Data and Meta-Data Acquisition}

We augmented tissue, sex and age based on human sRNA-seq expression profiles.
We used sRNA-Seq data data from the SEA \citep{Rahman17} that contains 4243 samples and annotations in 350 datasets.  The relatively large number of high-quality samples allowed us to use DL for data augmentation. Each sample contained annotations, sRNA expression counts of approximately 35000 sRNAs and expression information of potentially viral and bacterial transcripts (approximately 5600 'contaminants'), according to the output of the OASIS 2 sRNA analysis application \citep{Rahman18}. 
Tissue prediction was based on sRNA expression profiles only, because the use of contamination profiles did not noticeably increase the accuracy of prediction; for age and sex prediction, contamination profiles were used.
The number of datasets and samples is summarized in Table \ref{tab0}. 

\begin{table}[p!]
\caption{Number of samples and datasets used to augment tissue, age and sex. The samples comprised 42\% males and 58\% females.}\label{tab0}
\centering
\begin{tabular}{|c|c|c|}
\hline
{\bf Metadata field}&{\bf No. Datasets}&{\bf No. Samples}\\
\hline
Tissue&128&2806\\
Tissue after filtering&105&2215\\
Sex&41&1591\\
Age&27&888\\
\hline
\end{tabular}
\end{table}

To avoid small classes with specific tissues we 
%\subsection{Tissue grouping}
%\label{sec_tissuegr}
%Tissues are annotated using the 'most specific' terms available in SEA's Ontologies \citep{Rahman17}. For example, parts of the brain are annotated as "neocortex" or "prefrontal cortex" if this information is available. 
%This results in many specific classes
%The annotation of the data to very specific terms, however, results in many tissues with few samples, which results in class imbalances and impairs proper ML training. 
%To avoid this, 
merged the available tissues using BTO in the SEA (Table \ref{tab1}) according to Fiosina et al. \citep{Fiosina19}.% We also added 641 cell line samples with known tissue. %This grouping is in contrast to our work in Fiosina et al. \citep{Fiosina19}, where we classified non-grouped tissues.

\begin{table}[p!]
\caption{Tissue and cell line grouping according to ontologies.}\label{tab1}
\begin{tabular}{|p{0.2\textwidth}|p{0.8\textwidth}|}
\hline
{\bf Tissue group}&{\bf Contained tissues}\\
\hline
blood\_group & blood, blood plasma, blood serum, peripheral blood, umbilical cord blood, serum, buffy coat, immortal human B cell, liver, lymphoblastoid cell\\
brain\_group & brain, cingulate gyrus, motor cortex, prefrontal cortex, neocortex\\
epithelium\_group & skin, dermis, epidermis, breast, oral mucosa, larynx\\
gland\_group & prostate gland, testis, kidney, bladder, uterine endometrium, tonsil, lymph node\\
intestine\_group & intestine, colon, ileal mucosa\\
\hline
\end{tabular}
\end{table}

%subsection{Age grouping}

Age is a continuous variable and its exact prediction is a regression problem that might be highly inaccurate when solely based on sRNA expression information. To use the same methods as those used for the prediction of other annotation fields, we grouped ages into $k$ intervals, $k=2,3,4$. Table {\ref{tab1_1}} summarizes the intervals used for age prediction.

\begin{table}[p!]
\caption{Age intervals used in prediction.}\label{tab1_1}
\label{tab_numintervals}
\begin{tabular}{|p{0.3\textwidth}|p{0.3\textwidth}||p{0.3\textwidth}|}
\hline
{\bf 2 intervals}&{\bf 3 intervals}&{\bf 4 intervals}\\
\hline
[0;65],(65;110]&[0;45],(45;70],(70,110]&[0;30],(30;60],(60,80],(80,110]\\
\hline
\end{tabular}
\end{table}

\subsection{Data scaling and filtering}

%DL and RF models require different data pre-processing. 

\begin{figure}[h]
\centering
\includegraphics[width=0.9\textwidth]{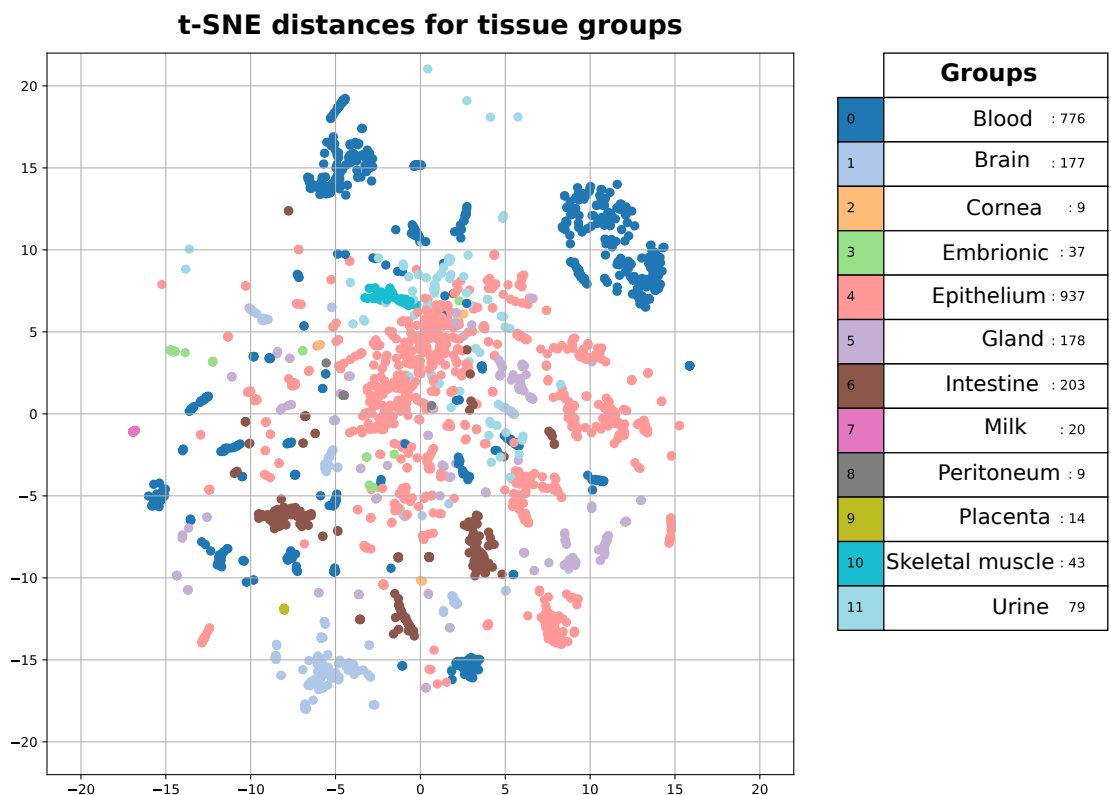}
\vspace{-0.7cm}
\caption{t-SNE Plot for available tissue types.} 
\label{fig1}
\end{figure}

Data scaling and filtering is described in detail in \citep{Fiosina19}. In brief, the counts were normalized using RPM, each factor was normalized using a MinMax Scaler, the factors containing more than 30\% zeros were removed (leaving aproximately 2500 sRNAs and 2000 contaminants). Small groups were also removed, leaving 105 datasets for the analysis (see Table \ref{tab0}). We observed 23\% cell lines and 77\% tissue samples in our data. Fig.~\ref{fig1} illustrates the t-SNE plot for the tissue groups after the sample filtering. 

\subsection{Models}
\subsubsection{DL model}

We used a fully connected neuronal network (NN) architecture. It comprised one input layer with the number of inputs equal to the number of variables after the initial filtering. The NN contained three hidden layers with 1000, 250, and 250 neurons and drop-out rates of 0.5, 0.4, and 0.4, respectively (achieved by a grid search). The number of neurons in the output layer was equal to the number of predicted classes. The output data (the annotation classes) was encoded with integers $(0, 1, 2, \ldots, num\_of\_classes)$, and transformed to categorical variables consisting of zeros and ones. We used the ReLU function to activate the input and the hidden layers, and the softmax function for the output layer, for a multi-class classification. As a loss function, either a binary or categorical cross-entropy was used. For the NN training, Adam optimizaer was used. The NN was trained for 50 epochs with a batch size of 30.

\subsubsection{RF model}
We used a two-stage classification strategy. First, we used all features (remained after filtering) for the classification and feature importance calculation. We used the top-1000 features rated by their Gini index for the second stage classification with an increased number of trees (500 instead of 100).
The mtry parameter was equal to the square root of the number of features. Given that the performance of RF method can be strongly affected by class imbalance, we down-sampled large classes to the size of the smaller ones.

\subsubsection{Validation:}
\label{validationSec}

We implemented two types of cross-validation (CV) to investigate the accuracy of the data augmentation. 
First, we used the average accuracy of 5-fold CV. In this scenario, the training cells and test samples were randomly selected a priori, so that in most cases, samples from each dataset (experiment) could be included in the training and test sets. 

%Second, we trained a model taking specific datasets out of the training set while they are added to the test set. We call this approach 'one dataset out' classification, a situation which reflects the real use case scenario in which a new dataset with unknown bias needs to be classified. More specifically, the dataset that is taike out of the training and included in the test set is of a class (e.g. tissue) that is included in the training set but from a different dataset (experiment). 

Then, we performed "one-dataset-out" classifications, where specific datasets were removed from the training set and incorporated into the test set after ensuring that the respective tissues still remained in the training data set.

Throughout this manuscript,  we refer to the 5-fold CV as "cross validation" and the validation for unseen datasets as "one dataset out".

\subsubsection{Deep Lift:}
\label{sec_deeplift_approach}

To biologically trace the decisions of the DL model to the input features, we used DeepLIFT scores. DeepLIFT \citep{Shrikumar17} is an approach to assign importance scores, which demonstrate how important the value of each particular input is for each particular output. The scores are assigned according to the difference between a given input and some reference (neutral) input. The DeepLIFT method overperforms other scoring methods \citep{Shrikumar17}; thus, it was selected for our analysis.
The DeepLIFT method calculates scores by backpropagating the contributions of all  neurons  in  the  network  to  every  feature of  the  input. 
Consequently, for each sample $i$, each input neuron $j$, and each output neuron $k$, a score $C_{i,j,k}$ is calculated, which represents an importance of an input $j$ for an output $k$ in the $i$-th input sample.
%(according to a reference input).

%\subsubsection{Explanation Approach:}

We have provided a three-step explanation of our augmentation models. 

Firstly, we used a heatmap to visualize the DeepLIFT scores of an individual sample. This enabled us to understand, which sRNAs are important for a particular prediction.

Secondly, we analyzed important sRNAs for each class $k$. We selected samples, which belonged to the class $k$: $y_i=k$ and calculated the average difference scores for 
the correct class and other classes: 
$$
D1_{j,k}=\avg\limits_{y_i=k} (C_{i,j,k} - \avg_{k' \ne k} C_{i,j,k'})
$$

Then, we selected the top $N$ sRNAs $j$ according to $D1_{j,k}$ for each class $k$.   

Finally, we investigated the number of sRNAs to be removed (to set their expression to zero), to change the classification results. For each sample $i$ of class $y_i=k$ and each class $k'\ne k$, we calculated the score differences
$$
D2_{i,j,k'} = C_{i,j,{y_i}}-C_{i,j,k'}.
$$

We ordered the differences $D2(i,j,k')$ and set the expression of sRNAs $j$ 0. We stop the process when the classification changes from $k$ to $k'$ (similarity analysis) or to any other class $k''\ne k$ (stability analysis). The corresponding average number of steps was applied to a matrix, which demonstrated "stability" of class (or "class similarity").
As a reference input, we used a vector of zeros. 

\subsubsection{Software libraries:}
All the scripts for DL classification are developed in R based on the "keras" library. The RF models are also implemented in R, using the "randomForest" library. For quality metrics, we used R "caret" package. We used the Python 3.5 "sklearn.manifold" t-SNE library to build the t-SNE plots. DeepLIFT was implemented using the "deeplift" Python library version 0.6.9.0. 

\section{Results}

%We checked three main hypothesis formulated in Section \ref{sec1}. First, we investigated whether and how accurately the annotation fields may be predicted based on sRNA expression, contamination profiles or both combined. We analyzed this prediction for the cross-validation and one dataset out tasks taking into account experimental bias. Second, we proof that augmentation based on DL might outperform classical approaches, such as RF to predict annotation fields. Third, we interrogated if DL model can be explained using DeepLift scores and if important miRNAs are  biologically relevant.

\subsection{Tissue prediction}
\subsubsection{Tissue group prediction:}
\label{sec_tissuegrpred}
We aimed to predict the tissue class, i.e., grouped tissue (\ref{tab1}). Although grouping leads to a smaller number of classes, it increases the samples per class. This should reduce the problems due to class imbalances and the over-fitting of very small training classes. The prediction was based on sRNA expression profiles only.

\begin{figure}[h]
\includegraphics[width=0.7\textwidth]{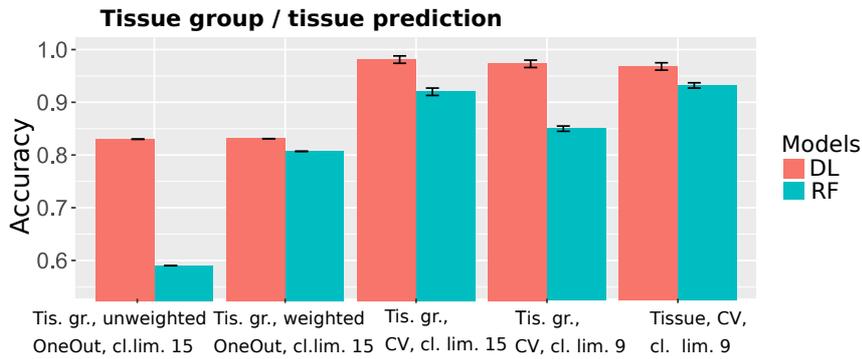}
\caption{CV and 'one dataset out' accuracy of tissues and tissue-groups}
\label{fig2}
\end{figure}

\begin{figure}[h]
    \begin{minipage}{0.5\textwidth}
\includegraphics[width=\textwidth]{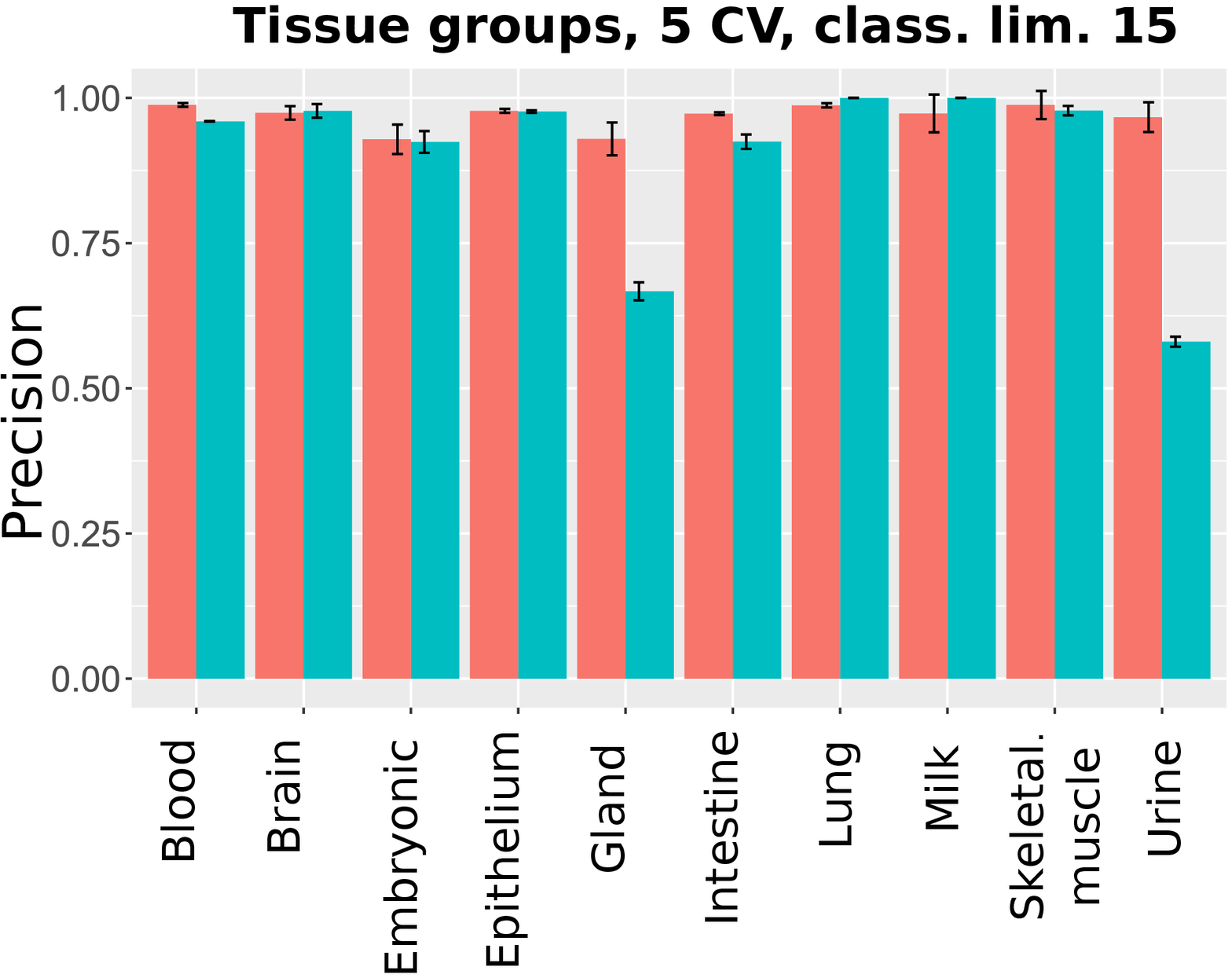}
    \end{minipage}\hfill
    \begin{minipage}{0.5\textwidth}
\includegraphics[width=\textwidth]{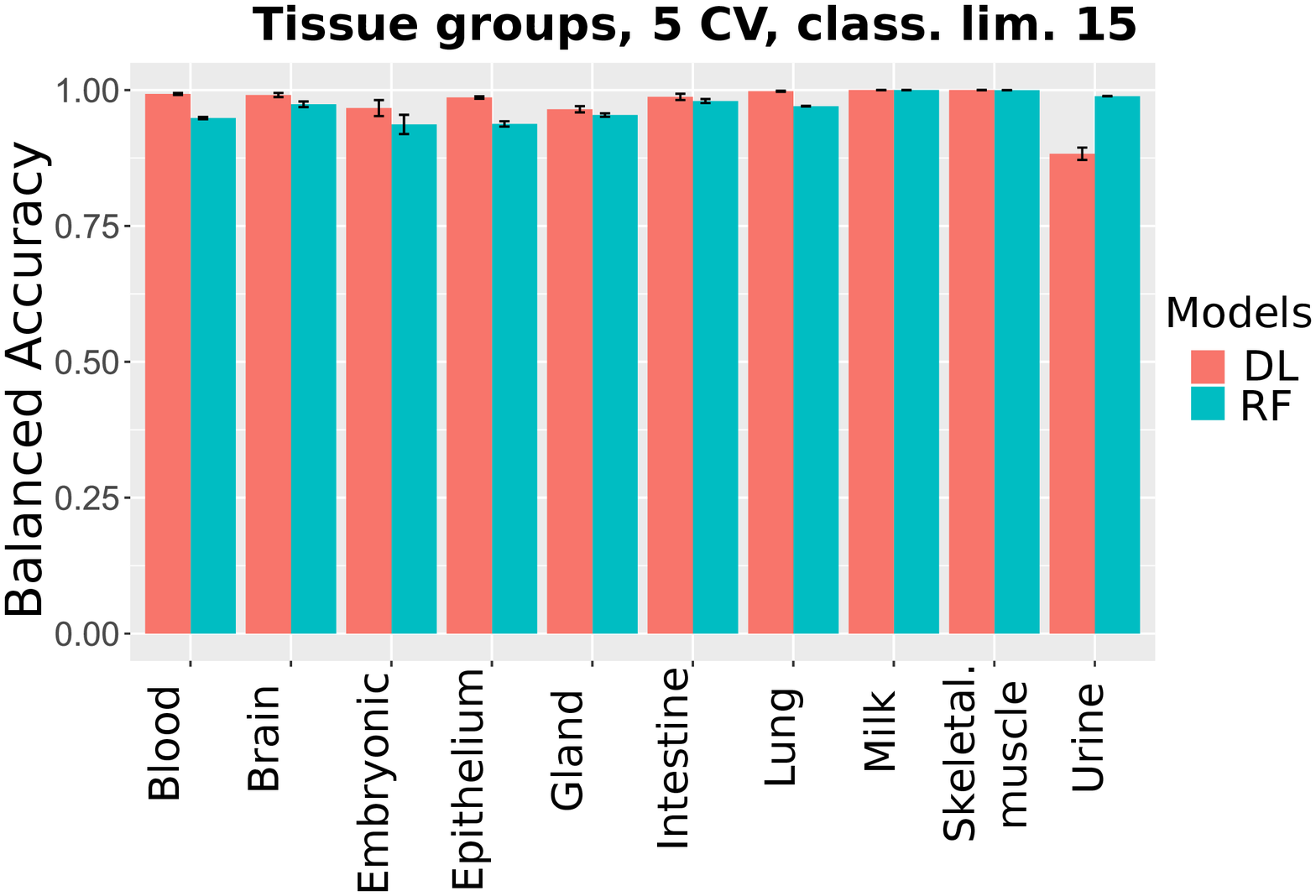}
    \end{minipage}\hfill
\caption{CV precision and accuracy for classes with min. 15 samples} 
\label{tissue_gr_cv_15}
\end{figure}

\begin{figure}[h]
    \begin{minipage}{0.5\textwidth}
\includegraphics[width=\textwidth]{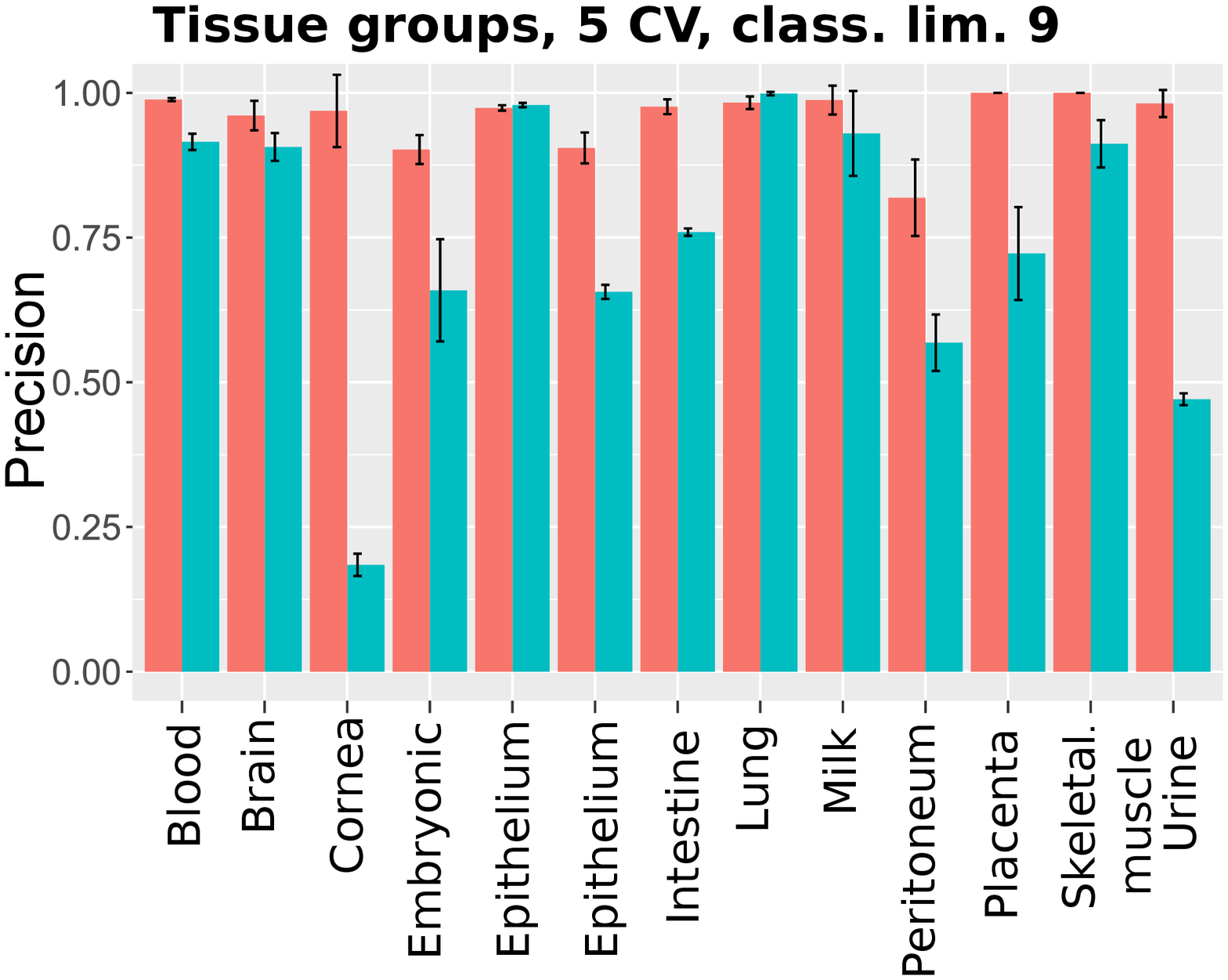}
    \end{minipage}\hfill
    \begin{minipage}{0.5\textwidth}
\includegraphics[width=\textwidth]{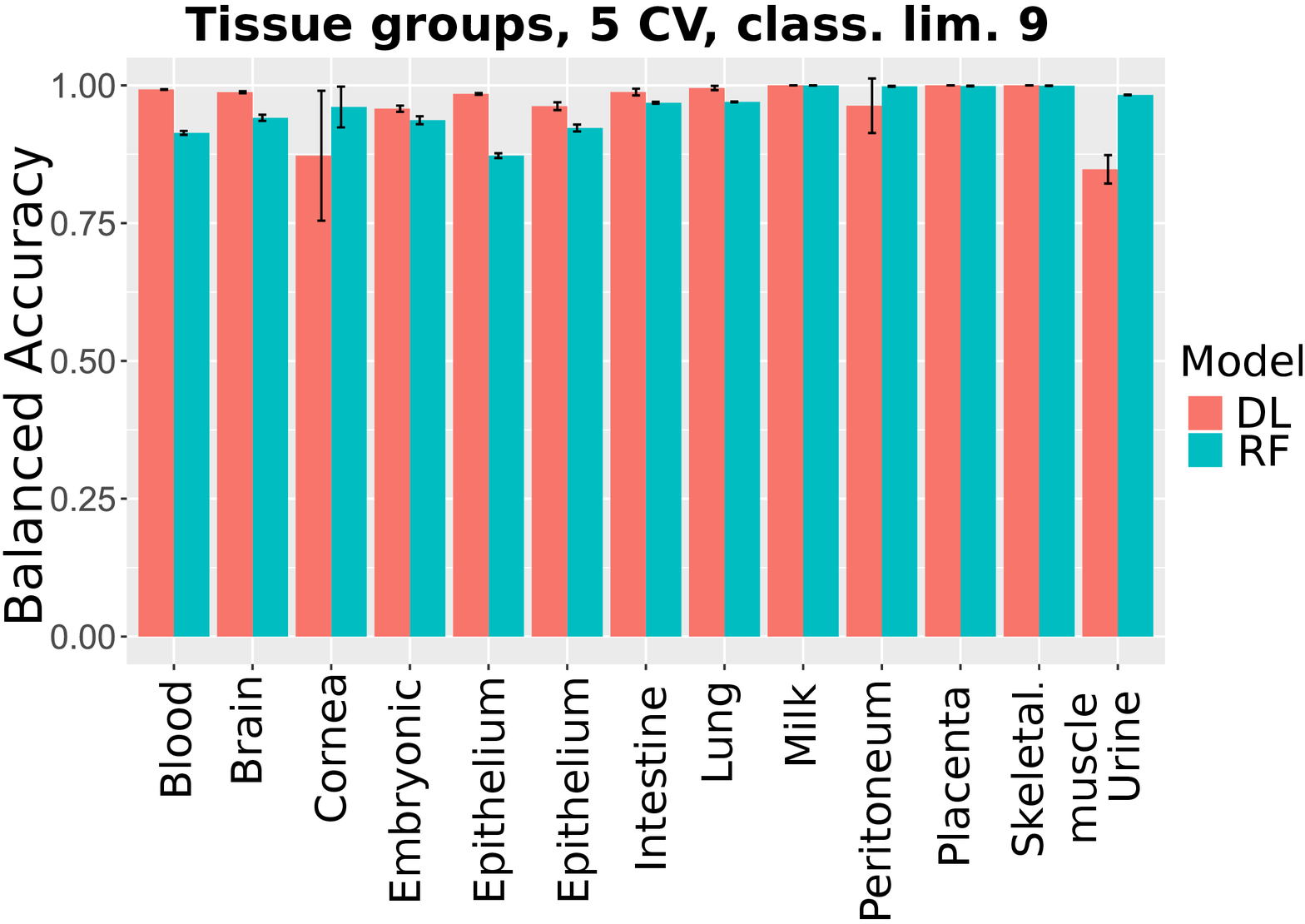}
    \end{minipage}\hfill
\caption{CV precision and accuracy for classes with min. 15 samples} 
\label{tissue_gr_cv_9}
\end{figure}

\paragraph{CV experiments}
In order to compare the performance of the DL and the RF models for datasets with a different degree of imbalance, we excluded classes for which less than 9 or 15 samples were available. Fig. \ref{fig2} shows that the RF model is less accurate, particularly for the threshold of 9 (DL: 97\%, RF: 85\%). For the threshold of  15, the accuracy increases, however it is still significantly inferior to that of the DL models (DL: 98\%, RF: 92\%). 
We surmised that the better performance of the DL model, together with the fact that the accuracy is only slightly affected by the minimum class size, can be attributed to its resilience to class imbalances (Fig. \ref{tissue_gr_cv_15} and \ref{tissue_gr_cv_9}).

%We surmise that DL model gave better results, in both cases, due to its resilience to class imbalances. Fig. \ref{tissue_gr_cv_15} and \ref{tissue_gr_cv_9} present precision and accuracy of CV models for each individual tissue group.

\begin{figure}[h]
    \begin{minipage}{0.5\textwidth}
\includegraphics[width=\textwidth]{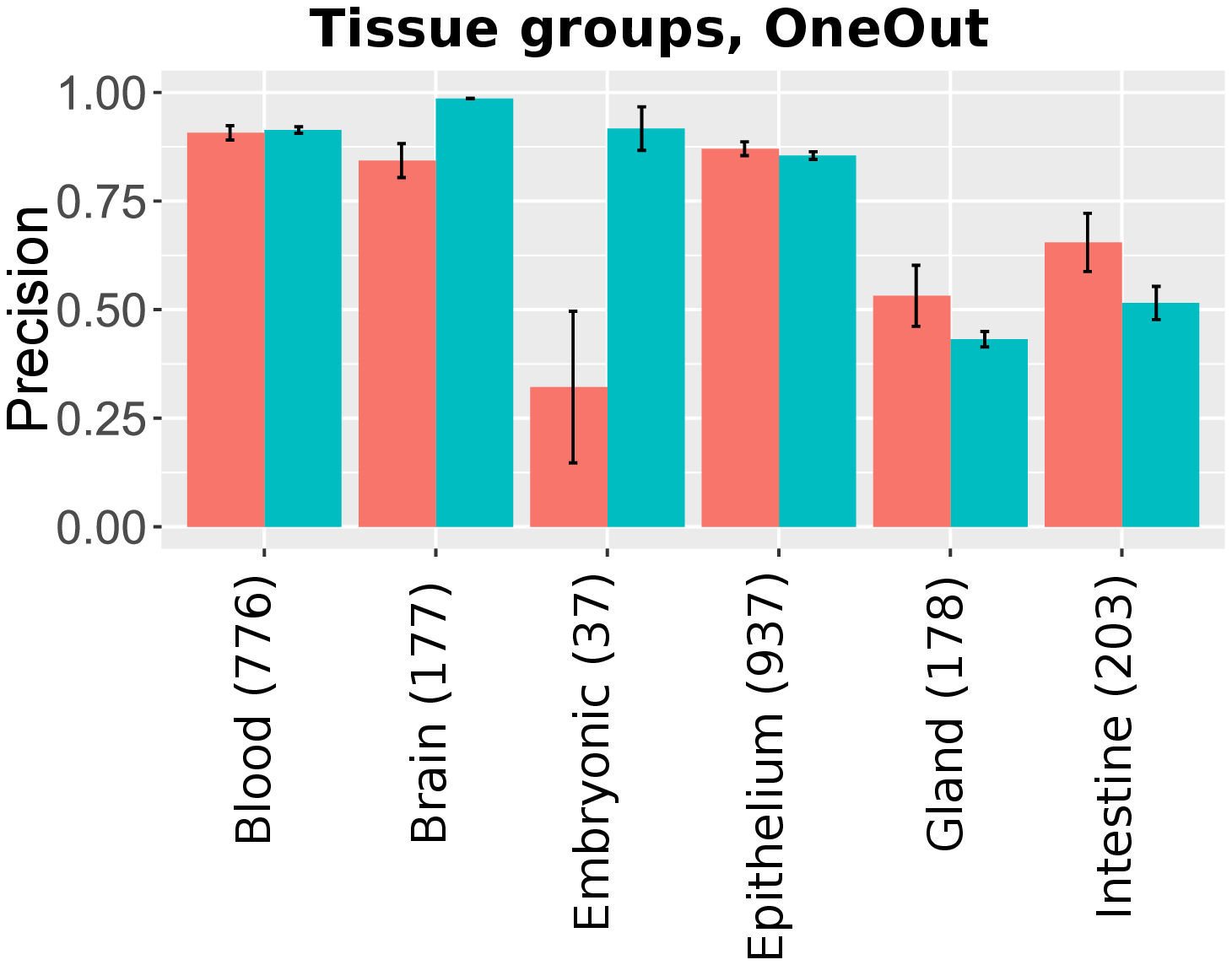}
    \end{minipage}\hfill
    \begin{minipage}{0.5\textwidth}
\includegraphics[width=\textwidth]{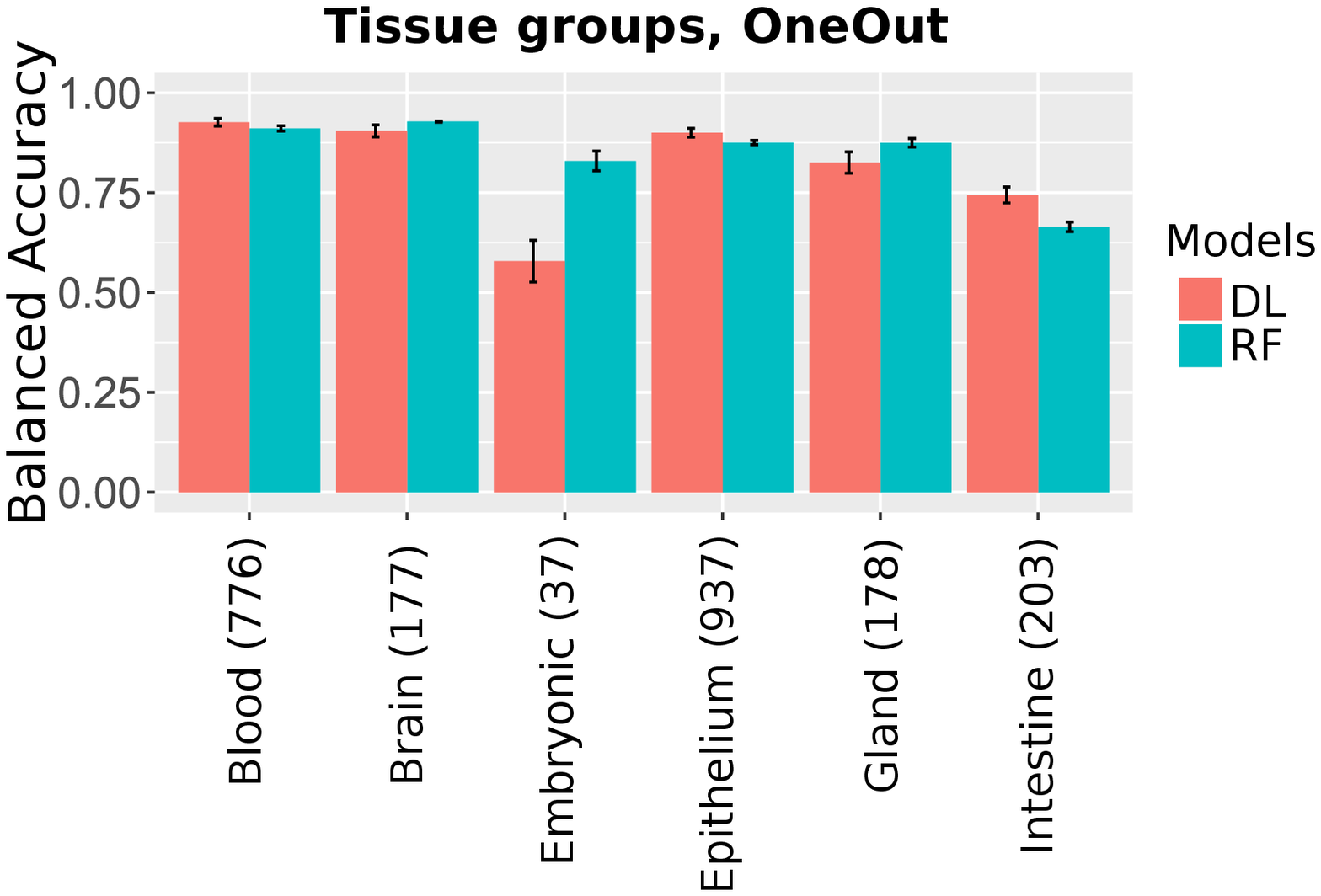}
    \end{minipage}\hfill
\caption{"One dataset out" tissue group prediction precision (left) and accuracy (right)} \label{fig_one_out_groups}
\end{figure}

\paragraph{"One dataset out" experiments} As detailed in the methods section, the aggregation of samples revealed 6 tissues with more than one dataset per tissue (see \citep{Fiosina19}, Sample Filtering). For the "one dataset out" classification, one dataset was removed from the training set and was only used for testing the classification accuracy, as can be observed from the last two columns of \ref{fig2}. This resembles a real augmentation scenario in which a dataset with an unknown bias is augmented by the ML algorithm. Although the datasets in the training and testing sets were derived from the same tissue, they, most probably, possessed very distinct biases that could have originate from varying library preparation methods, the biological conditions of the samples, cell types, and diseases.  The average accuracy of each group detection was 83.1\% (DL) and 80.7\% (RF) according to Fig. \ref{fig_one_out_groups}. 

\subsection{sRNA-seq sex prediction}\label{sexPredSec}

%Tissue prediction was based on sRNA profiles only. 
For the determination of sex, we opted for enlarging the data set with contamination expression counts. Effectively, we tried to predict sex with six different model: using sRNA expression counts, using contaminants, using both, each for the RF and the DL algorithm (Fig.\ref{fig51}). 
%To improve the model accuracy apart from sRNA-seq expression counts, we extended the models with contamination expression counts and compared sex prediction accuracy with and without contaminants (Fig.\ref{fig51}). 
The best models were the DL and RF models based on both sRNAs and contaminations, with an accuracy of 77\% and 76.9\%, respectively. The other three models RF(RNA), DL(contaminations), and DL(sRNAs) demonstrated an accuracy of approximately 76.2\%. It was unexpected that the model based only on contaminations predicted the sex with an accuracy of approximately 76\% for both DL and RF models. Thus, for sex prediction, the DL model outperformed the RF model slightly.

\subsection{sRNA-seq age prediction}
\begin{figure}[h]
    \begin{minipage}{0.6\textwidth}
\includegraphics[width=\textwidth]{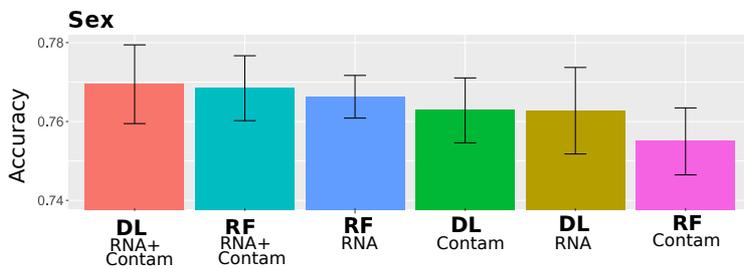}
    \end{minipage}\hfill
 
\caption{CV sex prediction accuracy with different models} \label{fig51}
\end{figure}

\begin{figure}[h]
   \begin{minipage}{0.5\textwidth}
\includegraphics[width=\textwidth]{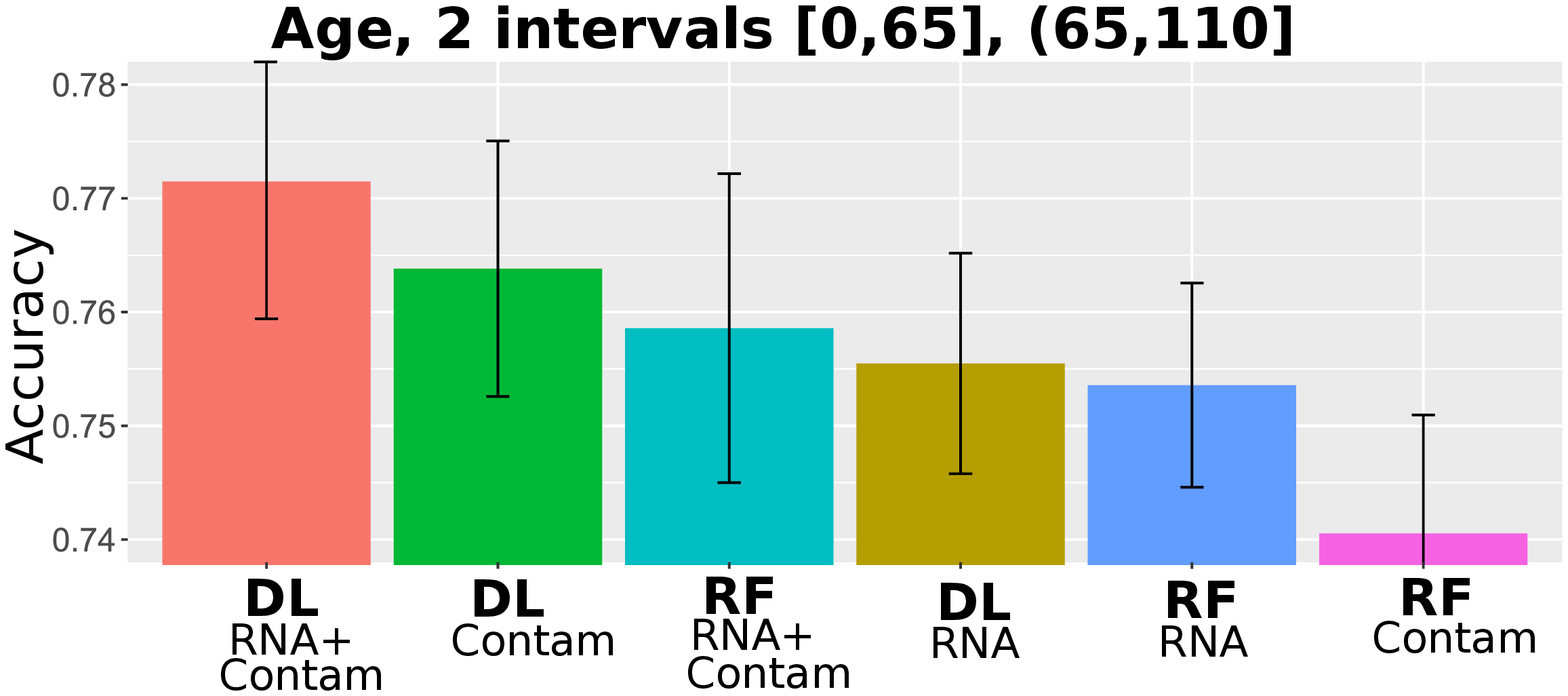}
    \end{minipage}\hfill
    \begin{minipage}{0.5\textwidth}
\includegraphics[width=\textwidth]{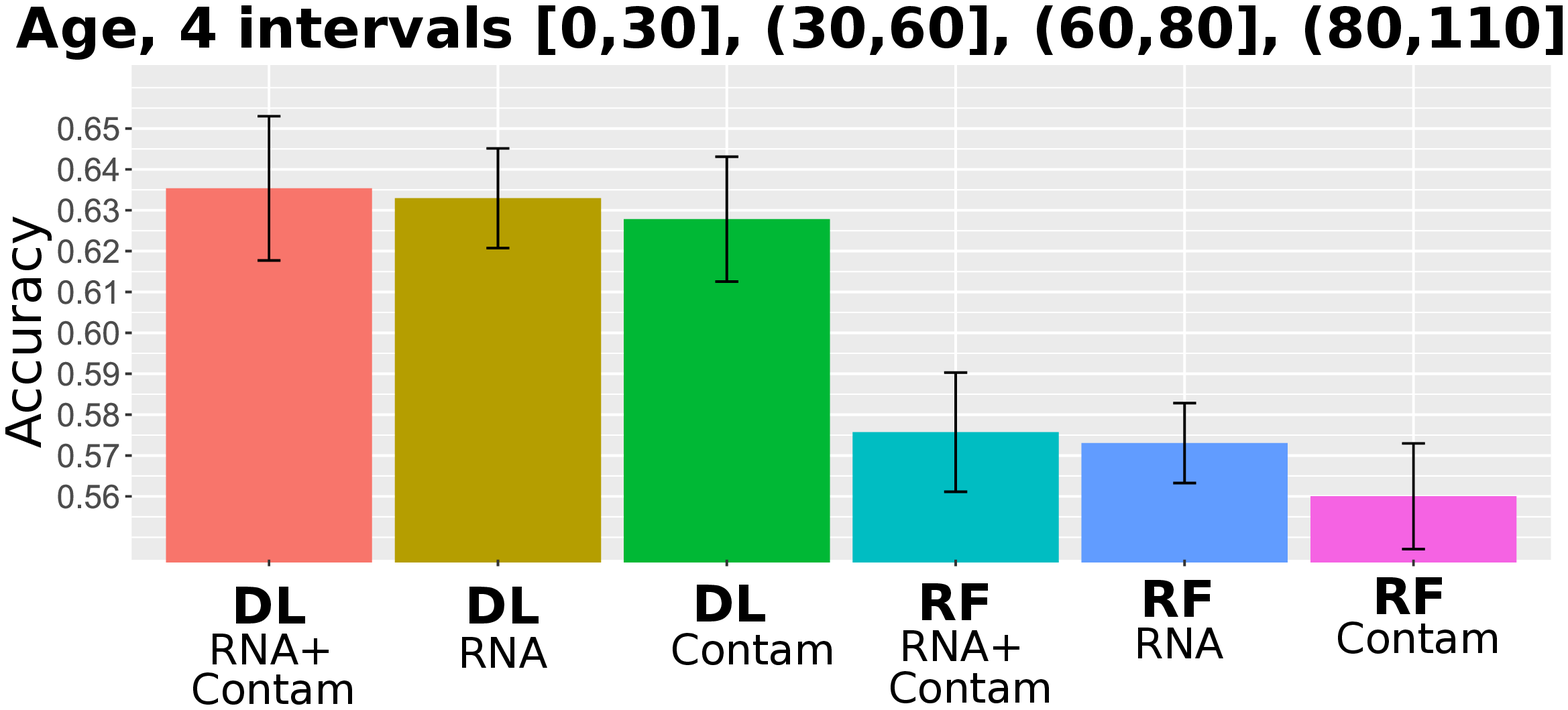}
    \end{minipage}\hfill
\caption{CV age prediction accuracy with different models} \label{fig52}
\end{figure}

For predicting the age, we used the contamination expression counts similar to sex prediction.
We predicted the age categories for three different splits yielding 2 and 4 categories, see Table \ref{tab_numintervals}. %Similar to sex prediction, we used the DL and RF models with standard parameters described above, as well as tried different models, which include  sRNA-seq expression counts and contamination expression counts. 
The results are presented in Fig. \ref{fig51} and \ref{fig52}. 

%The results demonstrate that training of a DL using sRNA and contamination counts gives the best accuracy (77.1\% for %2 intervals and 63.5\% for 4 intervals), followed by DL based either only sRNA or only contamination counts (76.4\% %for 2 intervals, and 63.2\% for 4 intervals). 
%RF demonstrated worse results (the best RF model gives 75.8\% for 2 intervals, and 57.5\% for 4 intervals).

A comparison of results using 2 and 4 categories demonstrates that accuracy decreases with increasing number of age categories for all models. In both cases, the DL models slightly outperform the RF model, in particular for a split into 4 intervals. Combined sRNA and contaminant data for a DL model yielded the highest accuracy, i.e. 77.1\% for a binary output and 63.5\% for 4 intervals. Notably, DL using less data (sRNA or contaminant data only) presents an accuracy of 76.4\% for binary output and 63.2\% for 4 intervals. The RF models performed slightly worse on average with a maximum accuracy 75.8\% for binary output and 4 intervals and 57.5\% for 4 intervals.

%In summary, DL slightly outperforms RF models in the augmentation of age and sex information.

\subsection{Explanation of deep learning results}
DL-based models are called "black boxes" because it is often unclear how the models arrive at their decisions. However, particularly in biological and medical settings, it is important to understand what enables algorithms to classify a sample, as the feature may be related to a cause as well as to a possible treatment. We investigated the explainability of automatic metadata augmentation with DL models, using backpropagation with the DeepLIFT method.

\subsubsection{Prediction explanation for individual samples}

\begin{figure}[h]
    \centering
\subfloat[Blood group sample]{{
\includegraphics[width=0.49\textwidth]{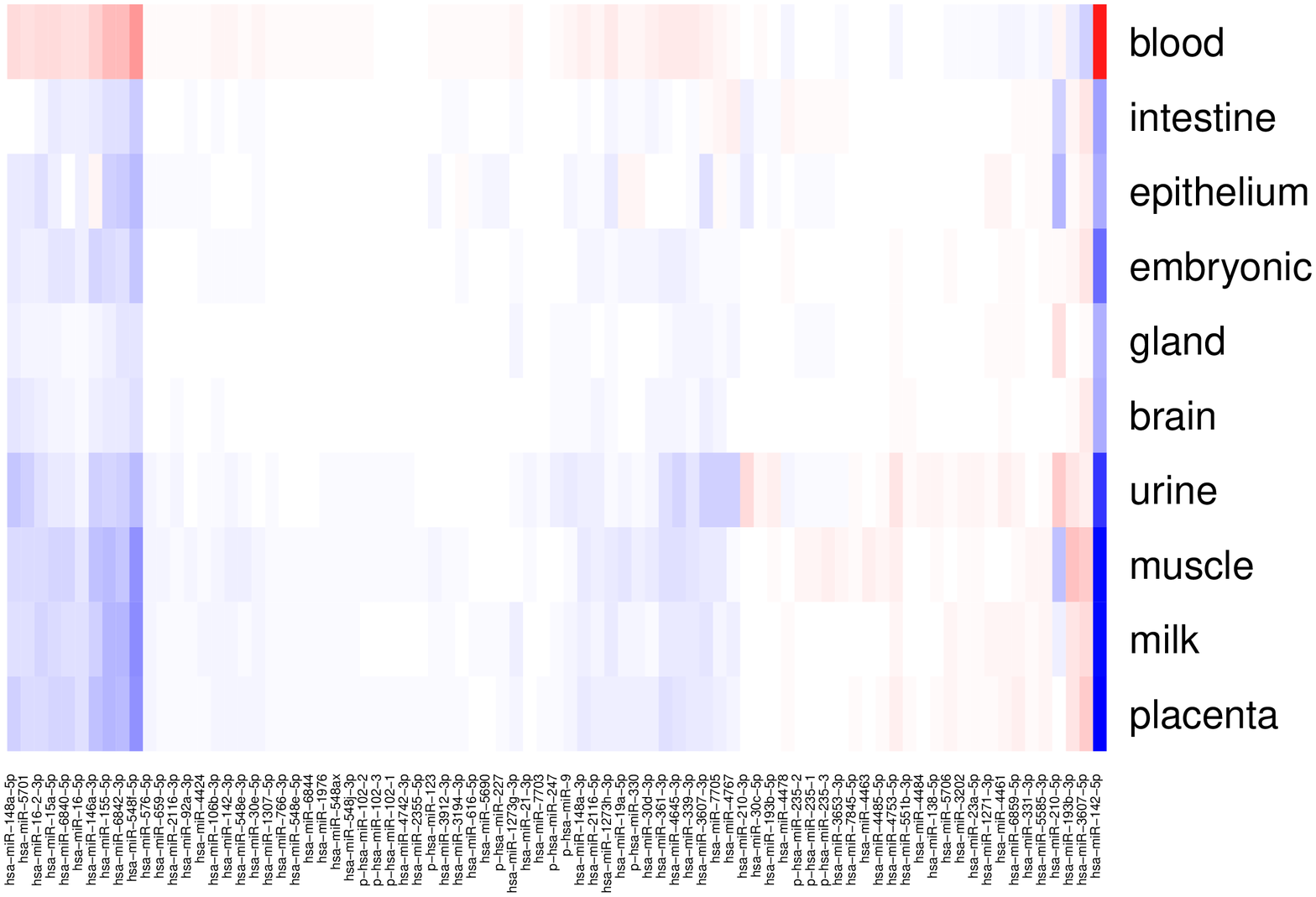}
}}
\subfloat[Brain group sample]{{
\includegraphics[width=0.49\textwidth]{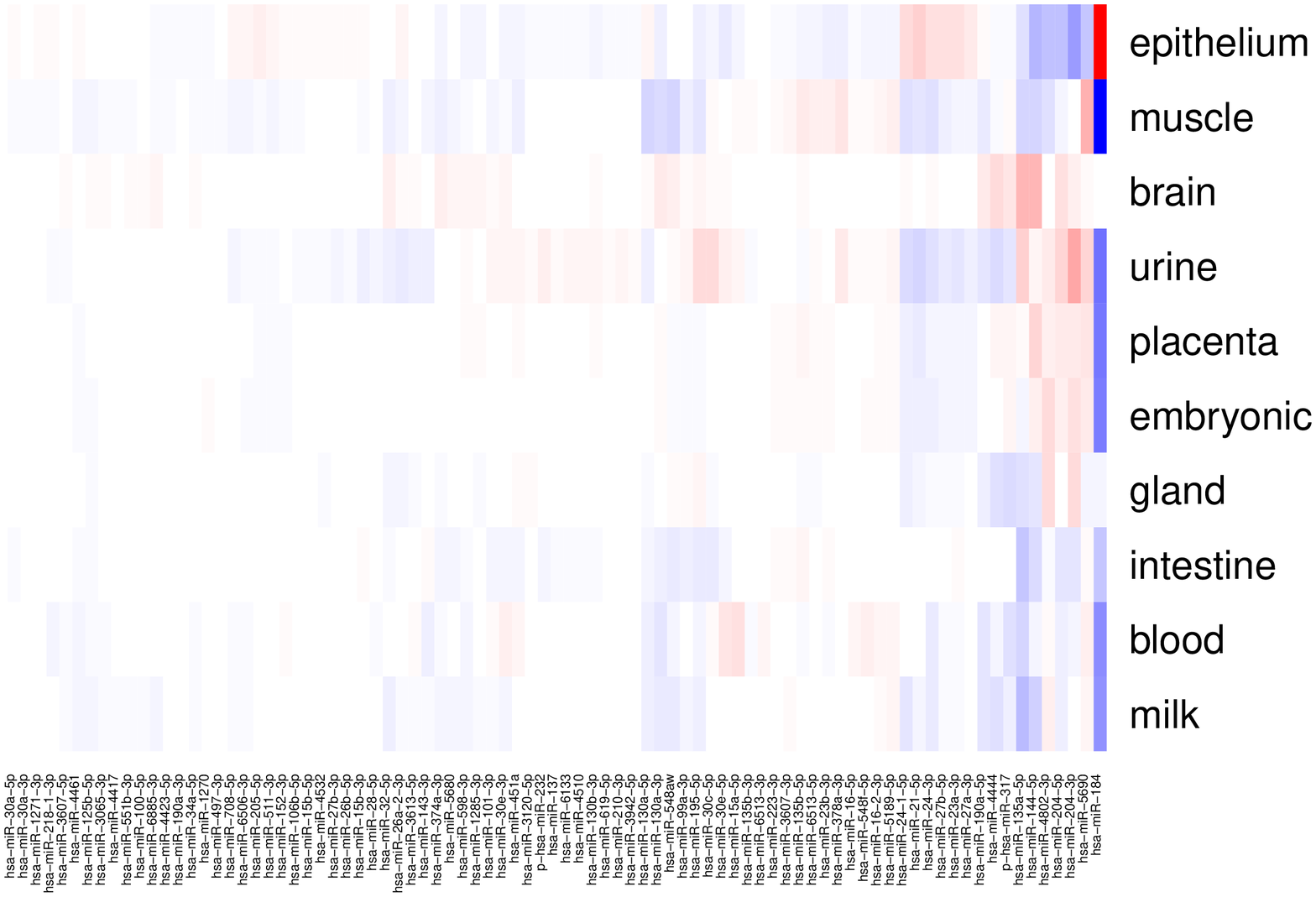}
}}
\caption{DeepLIFT scores for tissue group classification. On the left, we see some sRNAs clearly voting for the blood group (one sRNA on the right, and a cluster on the left). Particular sRNAs vote against the blood group (second and third on the left). Similarly, other tissues have specific sRNAs the score for or against the tissue.} \label{fig_tissue_scores}
\end{figure}

\begin{figure}[h]
    \centering
\subfloat[Female sample (correctly classified)]{{
\includegraphics[width=0.5\textwidth]{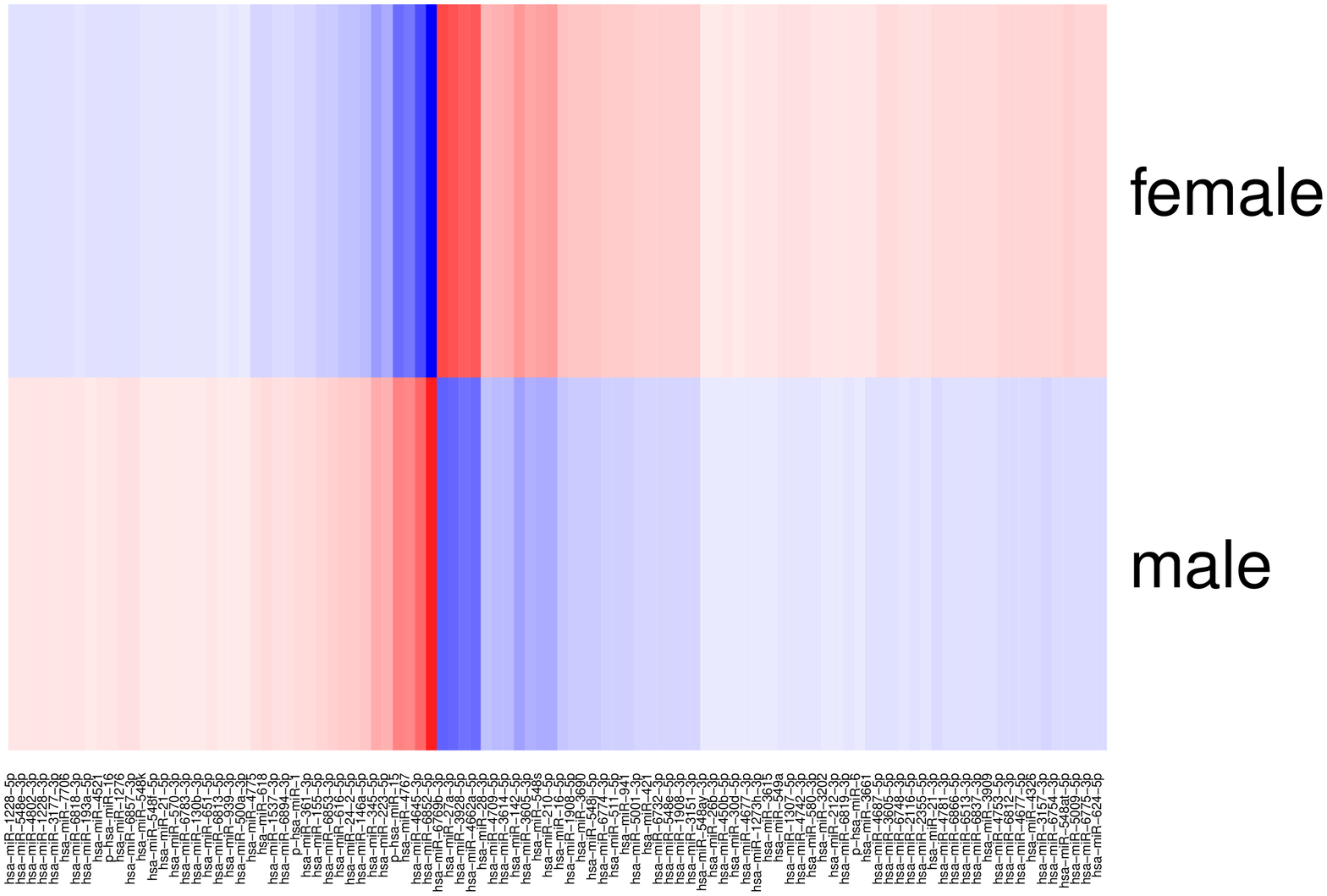}
}}
\subfloat[Female sample (incorrectly classified)]{{
\includegraphics[width=0.5\textwidth]{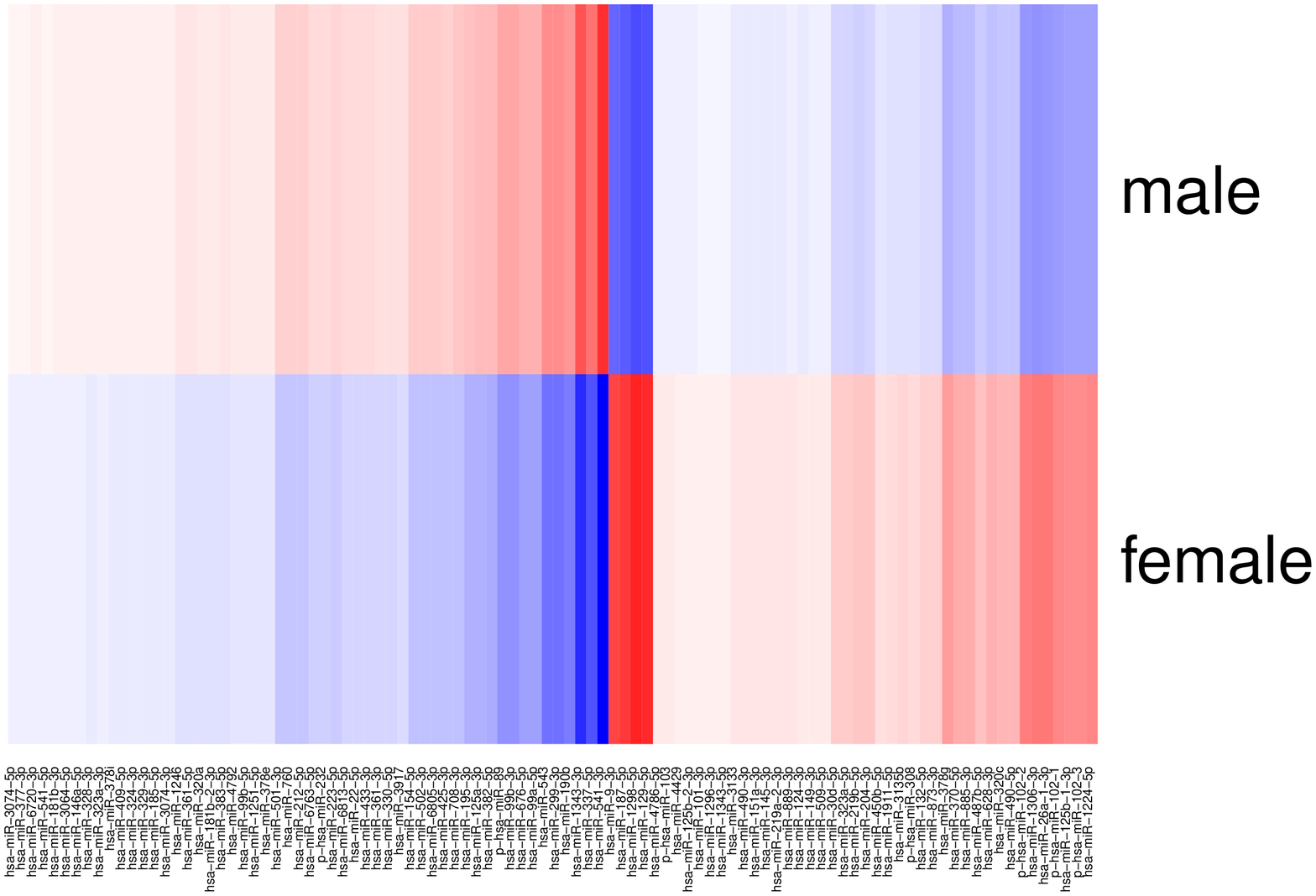}
}}
\caption{DeepLIFT scores for sex classification. For both samples, we see a number of sRNAs voting for and against each class. Both are classified as female. On the left, there is a majority of sRNAs voting for female.} \label{fig_sex_scores}
%\end{figure}

%\begin{figure}[h]
%    \centering
\subfloat[Old sample (correctly classified)]{{
\includegraphics[width=0.5\textwidth]{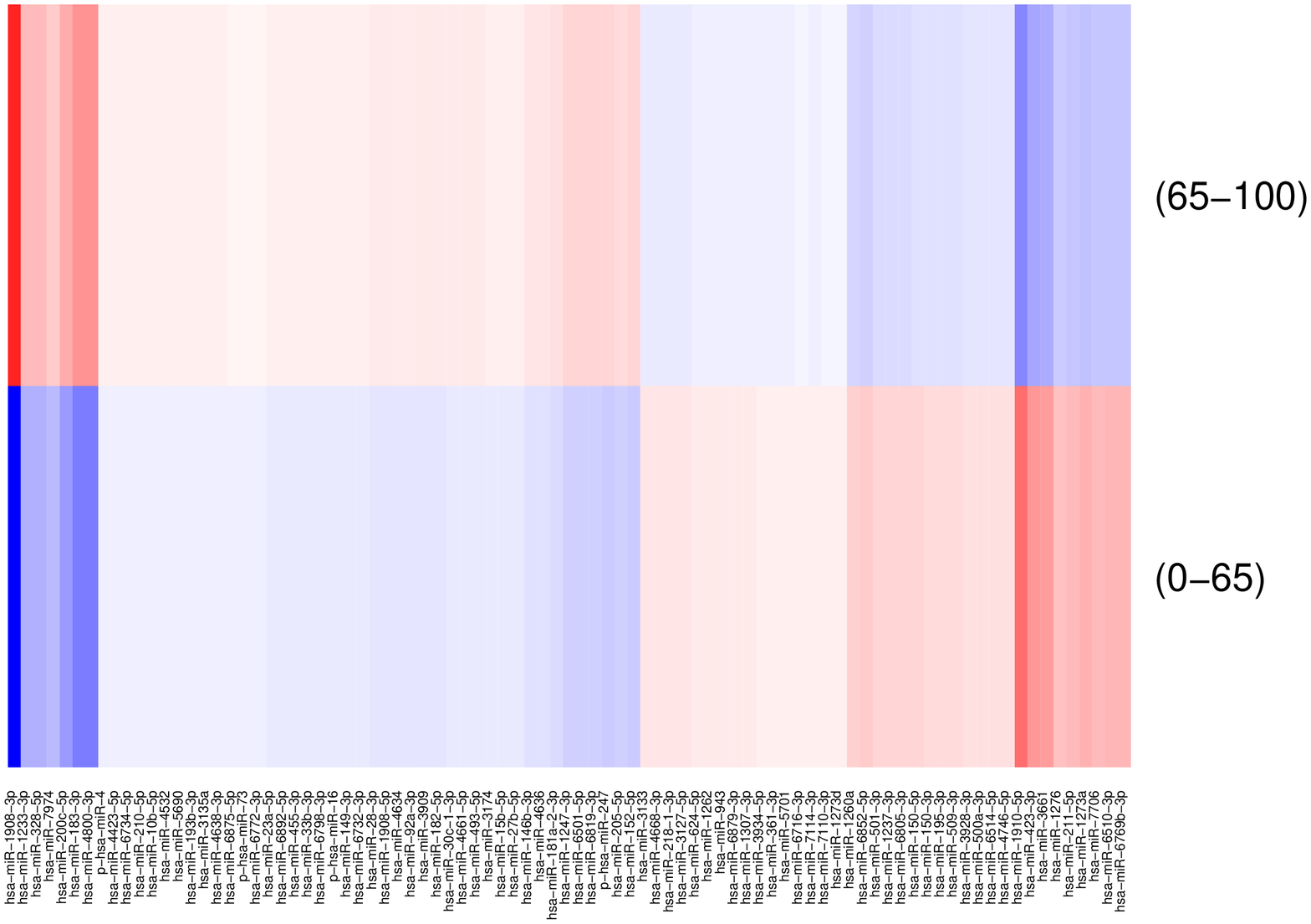}
}}
\subfloat[Young sample (correctly classified)]{{
\includegraphics[width=0.5\textwidth]{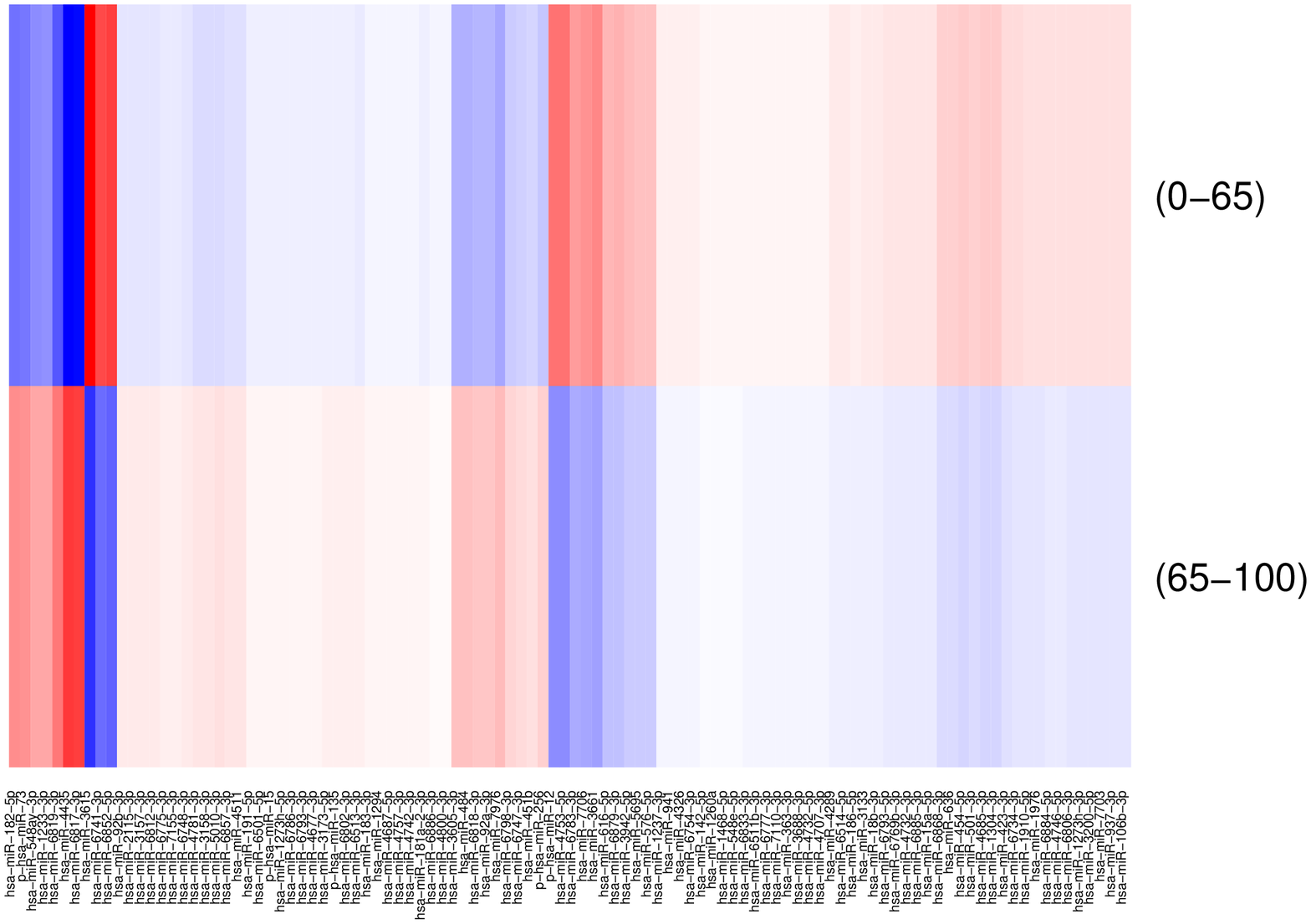}
}}
\caption{DeepLIFT scores for age classification. For both samples, we see a number of sRNAs voting for and against each class. However, the left sample has more sRNAs which vote for the old class and the right sample has more sRNAs that strongly vote for the young class.} \label{fig_age_scores}
\end{figure}

%We initially sought to understand how DeepLift scores explain a DL-based classification of individual samples. 
To visualize the backpropagation results, we used heatmaps that represented the scores for each individual class (Fig. \ref{fig_tissue_scores}, \ref{fig_sex_scores} and \ref{fig_age_scores}). 

The figures demonstrate that the visual representations of DeepLIFT scores may explain the factors important for a particular output of the NN.
%, as it is possible to visually discern sRNAs that are specific for a class or specifically absent. 
%We can see which sRNAs voted for the particular class and which were against. 
%This allows to overcome the disadvantage of DL that it could not be explained.
This shows that deep neural networks can indeed offer explainable and interpretable results.
\subsubsection{Average scores for sample prediction and enrichment}

%Explanation of individual sample prediction, however, does not allow to see the most important sRNAs for each class, because important sRNAs for prediction of individual samples change from sample to sample. So it is important to look to average scores among different samples.

%We calculated average importance scores $D1_{i,j,k}$ for each tissue class $k$ according to \ref{sec_deeplift_approach}. We combined all variables, sorted them by scores, and took the top $N=300$ sRNAs. For those sRNAs we calculated the average expression (Fig. \ref{tissue_heatmap}).

\begin{figure}[h]
 \centering
\subfloat[Tissue group prediction]{{
\includegraphics[width=0.5\textwidth]{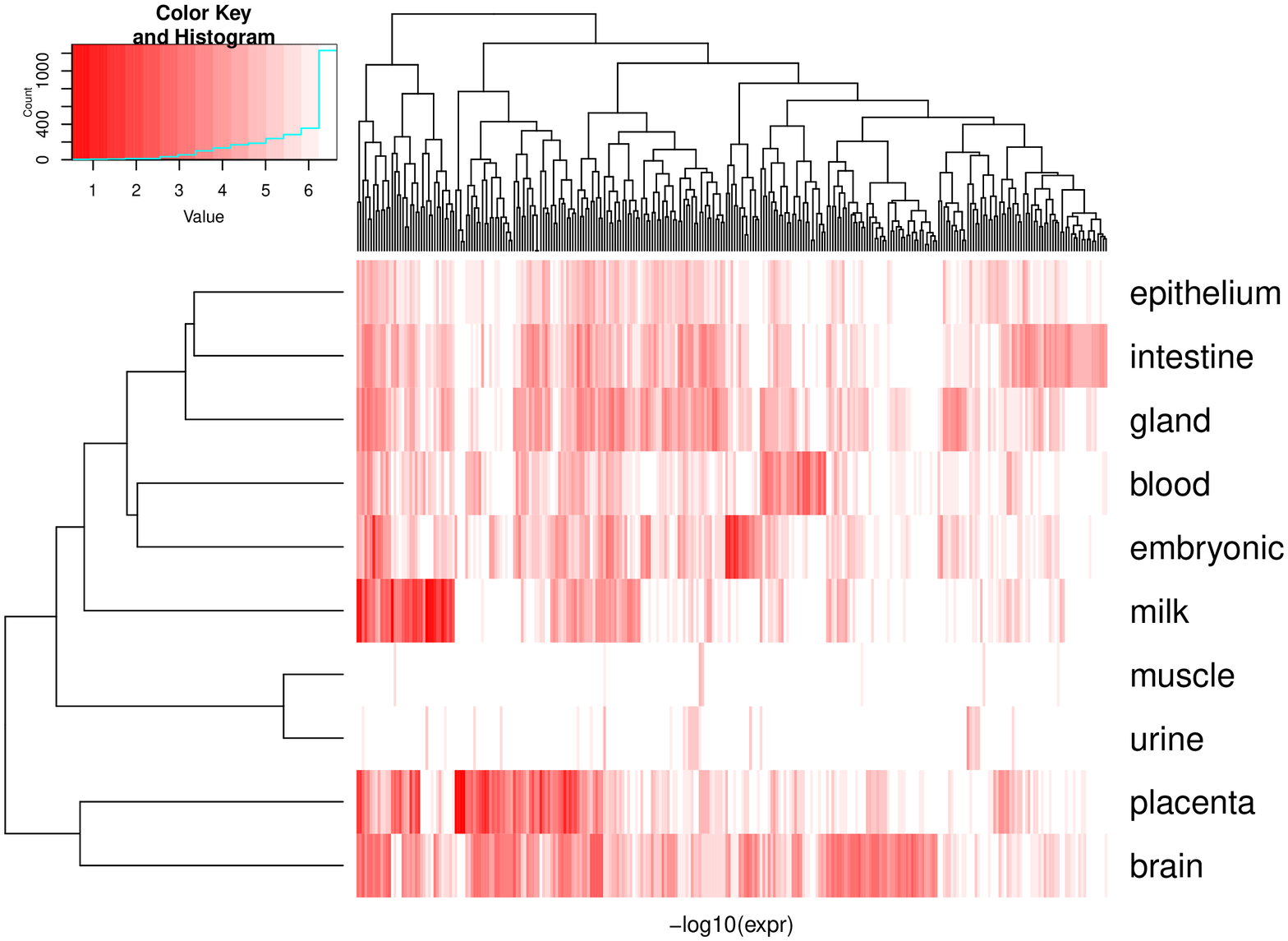}
}}
\subfloat[Sex prediction]{{
\includegraphics[width=0.5\textwidth]{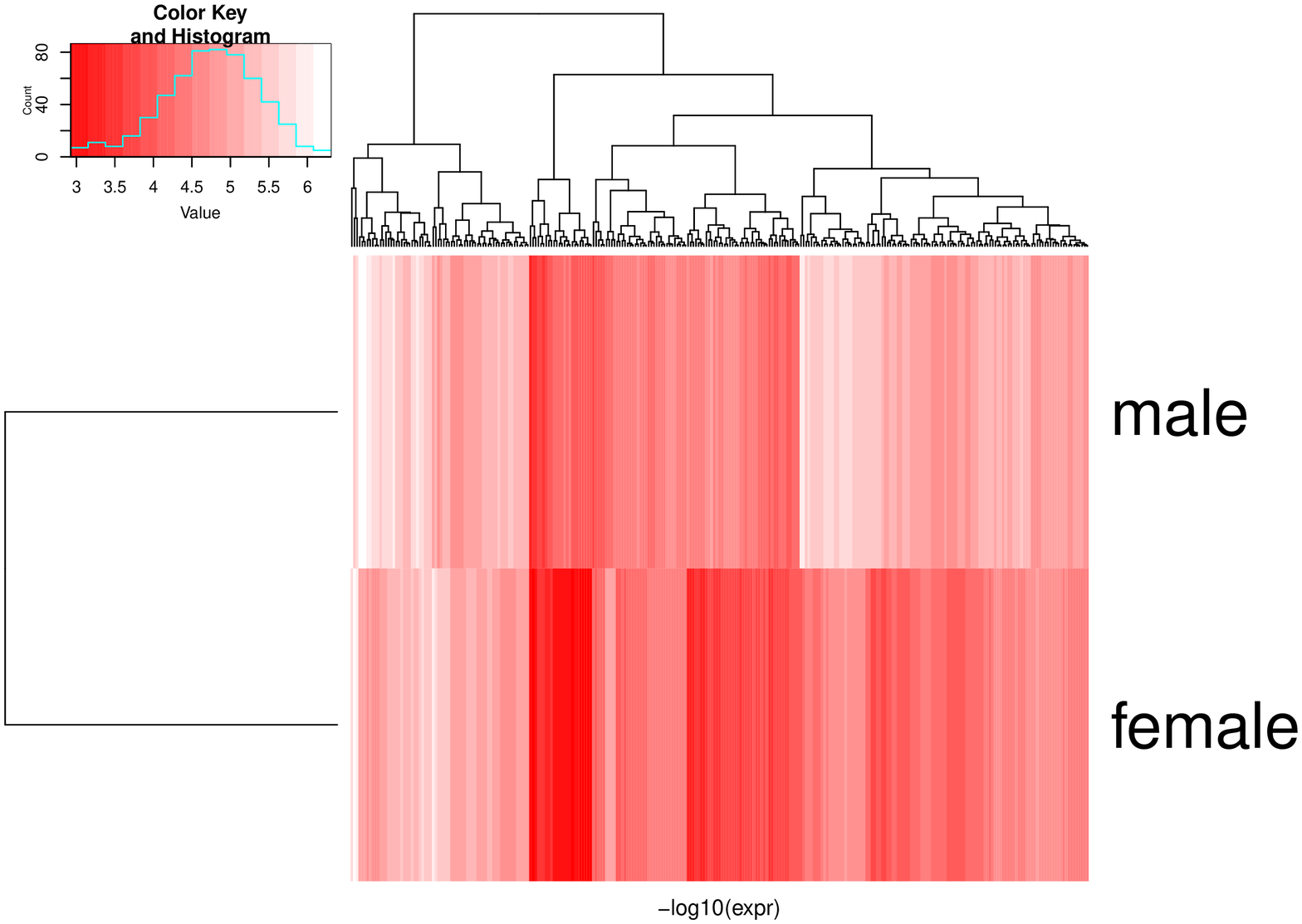}
}}
\caption{Expression of sRNAs with top 300 DeepLIFT scores.} 
\label{tissue_heatmap}
\end{figure}

The most important sRNAs for a class can, however, not be determined on a per-sample basis as individual samples show rather large variations. Thus, we computed $D1_{j,k}$  for each tissue $k$ as outlined in \ref{sec_deeplift_approach}. Using these scores, we selected the top $N=300$ sRNAs $j$ and calculated the average expression for each class (Fig \ref{tissue_heatmap}).

We observed that factors with big average DeepLIFT scores do not show a clear separation by expression levels. We still see some clusters of sRNAs, which are characteristic for the groups. These observations may be explained by non-linear class separation of the DL, which is not reflected just by average expression per class. 

\begin{figure}[h]
\includegraphics[width=\textwidth]{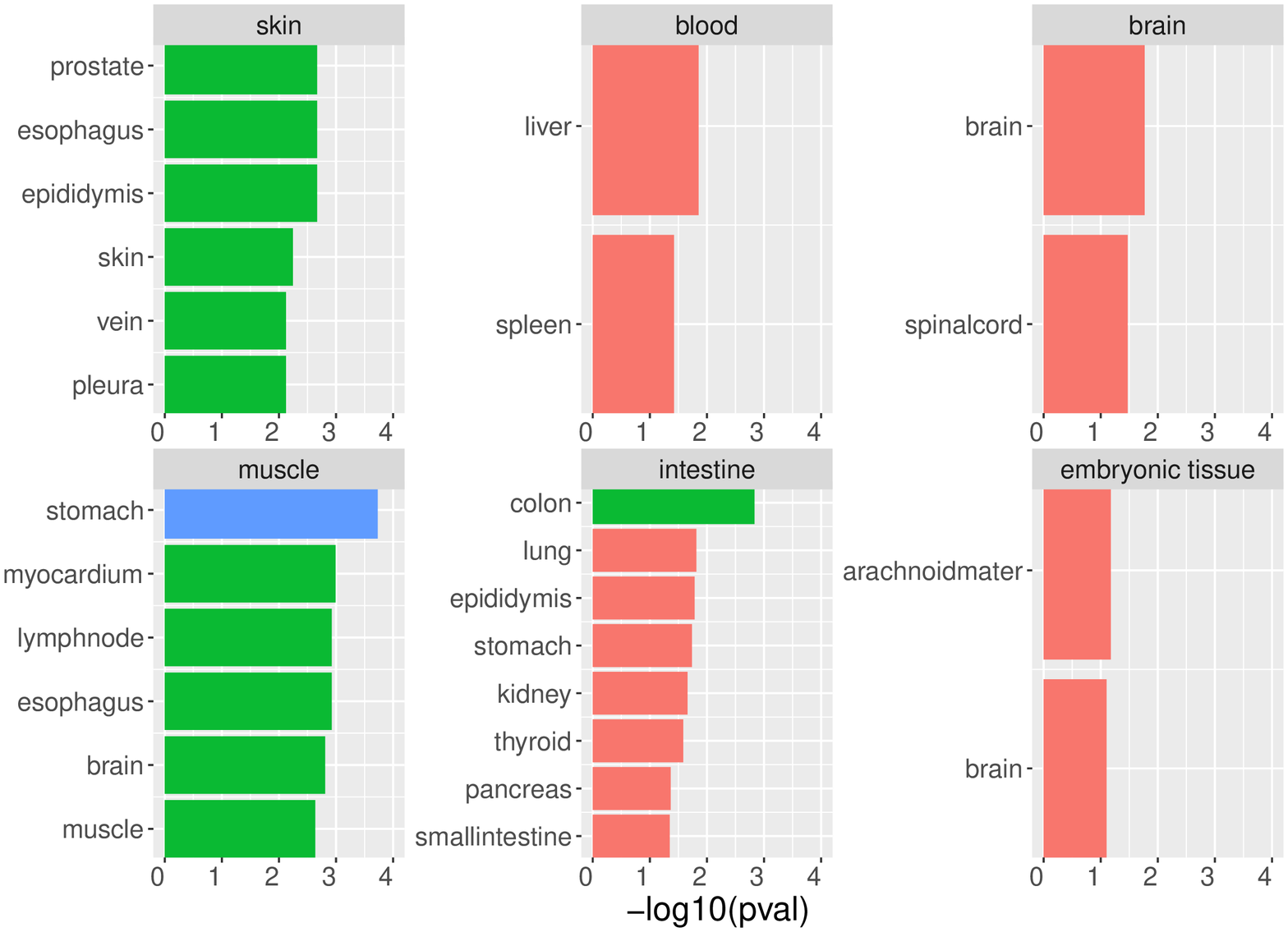}
\caption{Tissue enrichment. We see enrichment in 6 categories. For most of them, we see enrichment of tissue-relevant terms. For example, for epithelium we see skin, vein and pleura categories, for blood - liver and spleen, which are blood producing organs; for brain - brain and spinal cord, which are central nervous system relevant.} \label{fig_enrichment_tissues}
\end{figure}

To make sure that the results contain biologically relevant sRNAs, we investigated the enrichment of biological categories based on important sRNAs. We used the model based on miRNA only, as the enrichment information is available mostly for miRNAs (Fig. \ref{fig_enrichment_tissues} and Table \ref{tab_enrichment_tissues}).

%for miRNAs that are important for tissue classification.

\begin{table}[p!]
\caption{Enriched miRNAs for tissue prediction}\label{tab_enrichment_tissues}
\begin{tabular}{|p{0.2\textwidth}|p{0.8\textwidth}|}
\hline
{\bf Category}&{\bf enriched miRNAs}\\
\hline
skin & \footnotesize hsa-miR-205-5p; hsa-miR-205-3p; hsa-miR-193a-5p; hsa-miR-23a-3p; hsa-miR-21-5p; hsa-miR-3195; hsa-miR-27a-3p; hsa-miR-224-5p; hsa-miR-98-5p; hsa-miR-944
%; hsa-miR-200c-3p; hsa-miR-193a-3p; hsa-miR-24-3p; hsa-miR-24-1-5p; hsa-miR-324-5p; hsa-miR-125a-3p; hsa-miR-497-5p; hsa-miR-26b-5p; hsa-miR-106b-5p; hsa-miR-15b-5p; hsa-miR-99b-3p; hsa-miR-4649-3p; hsa-miR-374a-5p; hsa-miR-152-3p; hsa-miR-1237-3p; hsa-miR-141-3p; hsa-miR-708-5p; hsa-miR-361-5p; hsa-miR-365a-5p; hsa-miR-203a-3p; hsa-miR-4532; hsa-miR-6510-5p; hsa-miR-452-5p
\\
blood & \footnotesize hsa-miR-99a-5p; hsa-miR-142-5p; hsa-miR-4732-3p; hsa-miR-486-5p; hsa-miR-15a-5p; hsa-miR-1976; hsa-miR-16-5p; hsa-miR-16-2-3p; hsa-miR-129-5p; hsa-miR-1224-5p
%; hsa-miR-760; hsa-miR-636; hsa-miR-451b; hsa-miR-6513-3p; hsa-miR-3615; hsa-miR-1343-3p; hsa-miR-656-3p; hsa-miR-30e-5p; hsa-miR-92a-3p; hsa-miR-576-5p; hsa-miR-5189-5p; hsa-miR-432-5p; hsa-miR-484; hsa-miR-451a; hsa-miR-885-5p; hsa-miR-3605-3p; hsa-miR-361-3p; hsa-miR-326; hsa-miR-155-5p; hsa-miR-5010-3p; hsa-miR-181b-5p
\\
brain & \footnotesize hsa-miR-153-3p; hsa-miR-138-5p; hsa-miR-100-5p; hsa-miR-9-5p; hsa-miR-874-3p; hsa-miR-124-3p; hsa-miR-125b-5p; hsa-miR-181c-3p; hsa-miR-654-3p; hsa-miR-598-3p
%; hsa-miR-149-5p; hsa-miR-628-5p; hsa-miR-551b-3p; hsa-miR-1271-5p; hsa-miR-758-5p; hsa-miR-539-5p; hsa-miR-346; hsa-miR-338-5p; hsa-miR-195-5p; hsa-miR-431-3p; hsa-miR-181c-5p; hsa-miR-1185-5p; hsa-miR-488-3p; hsa-miR-328-3p; hsa-miR-139-5p; hsa-miR-3200-3p; hsa-miR-151a-5p; hsa-miR-219a-2-3p; hsa-miR-381-3p; hsa-miR-125b-2-3p; hsa-miR-487b-3p; hsa-miR-125b-1-3p; hsa-miR-137
\\
muscle & \footnotesize hsa-miR-378a-5p; hsa-miR-133a-3p; hsa-miR-193b-3p; hsa-miR-4463; hsa-miR-6723-5p; hsa-miR-4644; hsa-miR-1271-5p; hsa-miR-378a-3p; hsa-miR-4485-3p; hsa-miR-193b-5p
%; hsa-miR-6511a-3p; hsa-miR-4640-3p; hsa-miR-6511b-3p; hsa-miR-486-5p; hsa-miR-885-5p; hsa-miR-3653-3p; hsa-miR-6087; hsa-miR-5096; hsa-miR-1296-5p; hsa-miR-30d-5p; hsa-miR-2392; hsa-miR-331-3p; hsa-miR-197-3p; hsa-miR-3960; hsa-miR-23b-3p; hsa-miR-3620-5p; hsa-miR-628-3p; hsa-miR-328-3p
\\
intestine & \footnotesize hsa-miR-215-5p; hsa-miR-194-3p; hsa-miR-194-5p; hsa-miR-192-3p; hsa-miR-192-5p; hsa-miR-200b-3p; hsa-miR-200b-5p; hsa-miR-19b-3p; hsa-miR-31-5p; hsa-miR-200c-3p
%; hsa-miR-145-5p; hsa-miR-29a-5p; hsa-miR-3651; hsa-miR-135b-5p; hsa-miR-19a-3p; hsa-miR-223-3p; hsa-miR-3679-5p; hsa-miR-574-5p; hsa-miR-29b-3p; hsa-miR-940; hsa-miR-3653-3p; hsa-miR-548ai; hsa-miR-652-5p; hsa-miR-200a-5p; hsa-miR-23a-3p; hsa-miR-214-3p; hsa-miR-141-3p; hsa-miR-28-3p; hsa-miR-23b-3p; hsa-miR-200a-3p; hsa-miR-29b-1-5p
\\
embryonic tissue & \footnotesize hsa-miR-92b-3p; hsa-miR-18b-3p; hsa-miR-363-3p; hsa-miR-421; hsa-miR-3195; hsa-miR-335-3p; hsa-miR-887-3p; hsa-miR-3648; hsa-miR-4417; hsa-miR-130b-3p
%; hsa-miR-4749-5p; hsa-miR-4787-3p; hsa-miR-130a-3p; hsa-miR-20b-5p; hsa-miR-185-5p; hsa-miR-505-5p; hsa-miR-28-3p; hsa-miR-1275; hsa-miR-744-5p; hsa-miR-1247-3p; hsa-miR-3124-5p; hsa-miR-598-3p; hsa-miR-1270; hsa-miR-4458; hsa-miR-204-5p; hsa-miR-5010-5p
\\
\hline
\end{tabular}
\end{table}

%We can see that DeepLift scores allow to extract the most important sRNAs for a given class. 
Our enrichment analysis clearly shows an overrepresentation of biologically meaningful sRNAs for a given target tissue, demonstrating that DeepLIFT scores allow the extraction of important tissue-specific sRNAs. We conclude that DeepLIFT is a viable method to explain DL decisions for genomic data.

\subsubsection{Stability of Solution}

\begin{figure}[h]
    \centering
\subfloat[Class stability]{{
\includegraphics[width=0.5\textwidth]{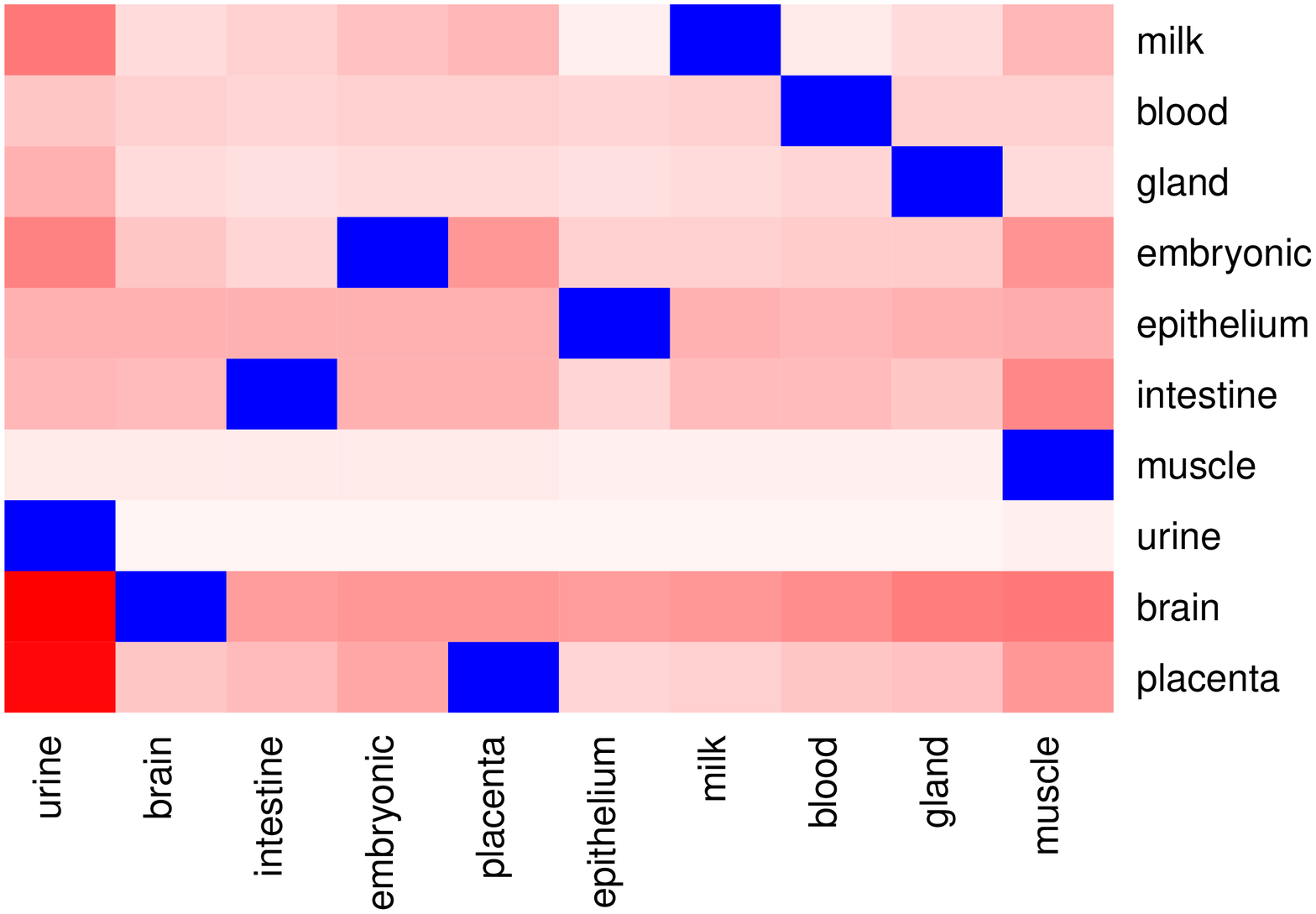}
}}
\subfloat[Class similarity]{{
\includegraphics[width=0.5\textwidth]{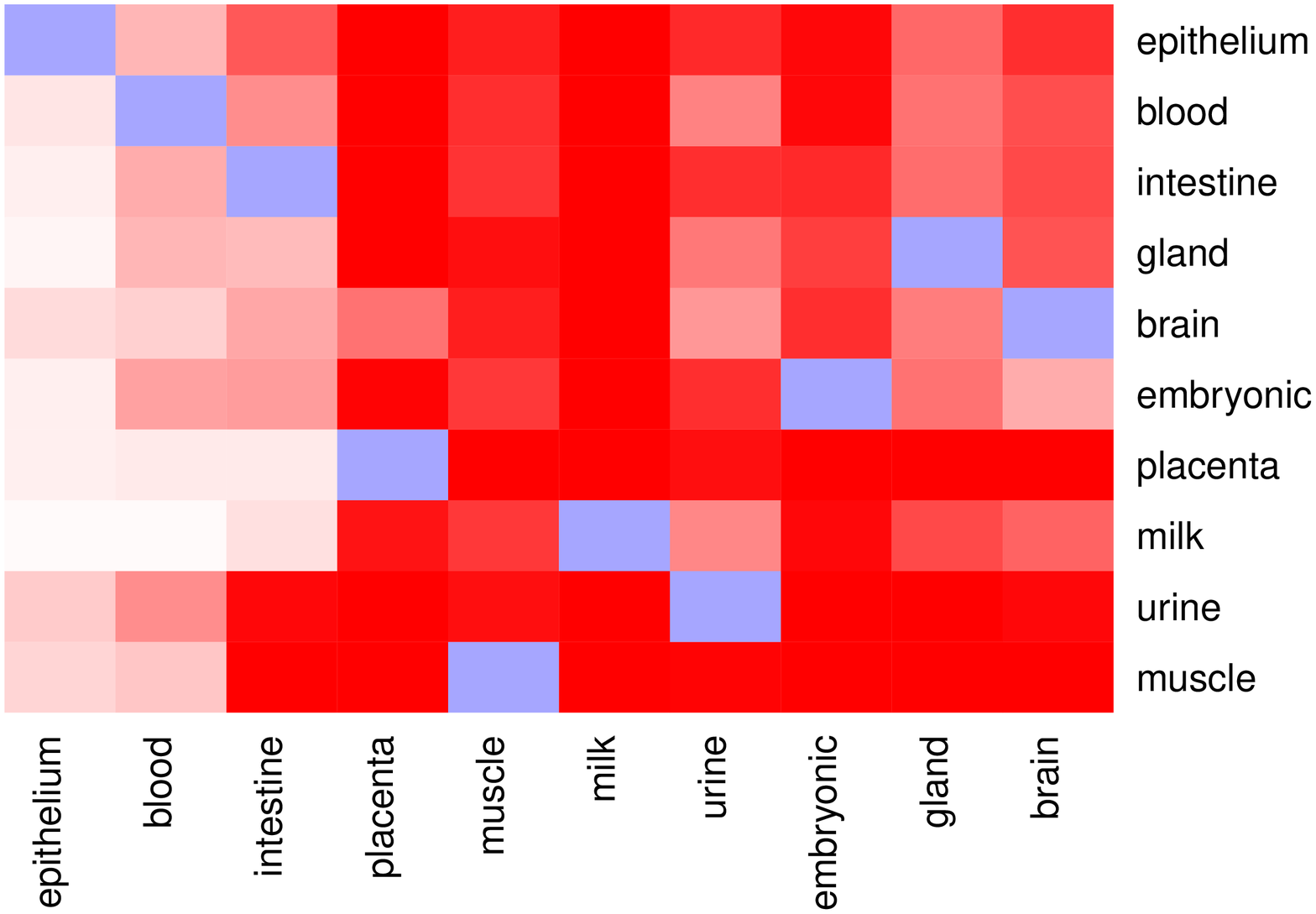}
}}
\caption{Stability and similarity of classes for tissue group classification. Intensity of red color shows number of steps to change class. Some classes are stable, such as the brain, intestine, and epithelium. Other classes like the urine, skeletal muscle, or gland are quite unstable and are easily transformed to the epithelium or blood groups.} \label{fig_stability}
\end{figure}

Further, we wanted to assess if DeepLIFT scores could provide insights into the stability of DL models. We ordered the differences $D2_{i,j,k'}$ and calculated the number of steps to change the predicted class according to \ref{sec_deeplift_approach} and (\ref{fig_stability}). Note that some classes are stable, other classes are quite unstable. These results most probably reflect the specificity and quantity of group-specific feature expressions. 
%We observe a very similar behavior for the classification of sex and age. 

\begin{table}[p!]
\caption{Enriched miRNAs for tissue prediction}\label{tab_stability_agegender}
\begin{tabular}{|p{0.22\textwidth}|p{0.1\textwidth}|p{0.1\textwidth}|p{0.22\textwidth}|p{0.22\textwidth}|}
\hline
{\bf Classification model}&{\bf Class 1}&{\bf Class 2}&{\bf No. of steps   Class1 into Class2}&{\bf No. of steps   Class2 into Class1}\\
\hline
Sex prediction&female&male&15.9&13.7\\
Age prediction&[0-65]&(65-110]&8.6&15.0\\
\hline
\end{tabular}
\end{table}

\section{Conclusion and future work}

Depending on the outcome variable (e.g. tissue, sex, age) automatic metadata augmentation can be a good option to annotate the missing metadata using sRNA expressions. The DL-based classification accuracy of tissue and sex predictions reaches 98\% and 77\%, respectively, the classification of age groups (or the regression of age) seems to demonstrate an inferior performance. 
In general, metadata augmentation, as undertaken in this study is dependent on the occurrence of the tissue of interest, or a similar tissue, in the training dataset of the classifier.
Another general problem is that of class imbalances and very rare classes. In this work, we have used an ontology-based grouping of rare classes to higher ontological nodes to increase the sample number for a given class. In future work, we plan to use a hierarchical classification, from general to specific tissue classes, to investigate the classification performance across the ontological hierarchy.
We have also demonstrated that, in general, the inclusion of contamination profiles in classification models improves the accuracy for sex and age.

sRNA expression profiles seem to be suitable for the augmentation of tissue information. A CV-based tissue group classification achieves  an accuracy over 98\%. 
In the "one dataset out" scenario, with a specific data set with a specific bias missing from the training data, samples from the ”unseen” dataset are classified with an accuracy of approximately 80\%.

For sex classification, the DL model achieved an accuracy of about 77\%, which may not be sufficient for accurate sex classification. This relatively low accuracy indicates that there may be no sex-specific expressed sRNAs for the X- or Y-chromosomes. 
Similarly, we obtained an accuracy of 77\% for predicting whether a person is younger or older than 65. For a split into 4 intervals, accuracy decreased to 64\%, indicating that the sRNA transcriptome does not consistently change with .
 
Lastly, we demonstrated that DL models can be explained both for individual samples and on average. For this purpose, the DeepLIFT scores demonstrated very promising results. 

\section*{Acknowledgments}

We would like to thank Daniel Sumner Magruder and Hannes Wartmann for helpful and essential suggestions.
The  research was supported by the  German  Federal  Ministry  of  Education  and Research (BMBF), 
project Integrative Data Semantics for Neurodegenerative research (031L0029);  
by German Research Foundation (DFG), project Quantitative Synaptology (SFB 1286 Z2) and by Volkswagen Foundation.

%\small

\section*{References}

\bibliography{FiosinaFiosins_ISBRA}
\end{document}